\documentclass{article}

\usepackage{longcat_style}
\usepackage{adjustbox}
\usepackage[utf8]{inputenc} %
\usepackage[T1]{fontenc}    %
\usepackage{newunicodechar}
\usepackage{hyperref}       %
\usepackage{xcolor}
\usepackage[normalem]{ulem} %
\hypersetup{
    colorlinks=true,      %
    linkcolor=mygreen,      %
    urlcolor=blue,       %
    citecolor=mygreen,      %
    linkbordercolor=blue, %
    urlbordercolor=blue,
    citebordercolor=blue,
    pdfborderstyle={/S/U/W 1}, %
}
\usepackage{float}
\usepackage{url}            %
\usepackage{booktabs}       %
\usepackage{amsfonts}       %
\usepackage{nicefrac}       %
\usepackage{microtype}      %
\usepackage{lipsum}		%
\usepackage{graphicx}
\usepackage{natbib}
\usepackage{doi}
\usepackage{amsmath}
\usepackage{amssymb} %
\usepackage{xspace}
\usepackage{enumitem}
\usepackage{multirow}
\usepackage{subcaption} 
\usepackage{makecell}
\usepackage{hyperref, cleveref}
\usepackage{pifont}
\usepackage[inkscapelatex=false]{svg}
\usepackage{subcaption}
\usepackage[hang, flushmargin]{footmisc}
\usepackage{wrapfig}
\usepackage{caption}
\usepackage[flushleft]{threeparttable}
\usepackage[most]{tcolorbox}
\usepackage{titletoc}
\tcbuselibrary{listings}

\setlist[itemize]{leftmargin=*}
\setlist[enumerate]{leftmargin=*}
\setlist[description]{leftmargin=*}

\usepackage{colortbl}
\definecolor{mygray}{gray}{.88}
\definecolor{mycyan}{cmyk}{.15,0,0,0}
\definecolor{mycyan2}{cmyk}{.85,0,0,0}

\definecolor{mygreen}{rgb}{0.19, 0.79, 0.02}
\definecolor{midnightgreen}{rgb}{0.0, 0.29, 0.33}

\definecolor{linegray}{gray}{.70}
\definecolor{linepurple}{rgb}{0.251, 0.043, 0.627}

\usepackage{bbm}              
\usepackage{amsthm}           

\title{Distilling Temporal Search and Reasoning: Evolving LLMs for Future Prediction via Harness-Assisted Efficient Data Synthesis}

\author{
    \textbf{Wanxu Cai}\textsuperscript{1,2,$\dagger$}, 
    \textbf{ Zhengyu Chen}\textsuperscript{2,\S}, 
    \textbf{ Huaisheng Zhu}\textsuperscript{2}, 
    \textbf{ Wei Wang}\textsuperscript{2}, 
    \textbf{ Jingang Wang}\textsuperscript{2}, 
    \textbf{ Qiang Xu}\textsuperscript{1,\S}
    \\
    \textsuperscript{1}The Chinese University of Hong Kong \\
    \textsuperscript{2}Meituan LongCat Team
}

\usepackage[most]{tcolorbox} 

\newtcblisting{guiguardprompt}[1]{%
  breakable,                    
  colframe=green!45!black,      
  colback=green!5!white,        
  colbacktitle=green!35!black,  
  coltitle=white,               
  fonttitle=\bfseries,          
  listing only,
  title={#1},                   
  listing options={
    basicstyle=\ttfamily\small,
    breaklines=true,            
    columns=fullflexible
  }
}

\begin{document}
\maketitle

\begingroup
\renewcommand{\thefootnote}{}
\footnotetext{\textsuperscript{$\dagger$}Work done during an internship at Meituan. Correspondence: \texttt{1155276131@link.cuhk.edu.hk}}
\footnotetext{\textsuperscript{\S}Corresponding author.}
\endgroup

\begin{abstract}
Future event prediction carries broad social impact yet remains challenging. SOTA approaches augment LLMs with external agent frameworks whose predictive capability vanishes once the harness is removed. While recent Tool-Integrated Reasoning (TIR) internalizes deep search for multi-hop retrieval of facts, forecasting further demands temporal search and reasoning over historical trends and dynamic shifts. The key obstacle is data: historical queries induce temporal leakage that degrades forecasting into retrieval. Prior works either freeze information gathering with static observations, or rely on rejection sampling or unresolved fresh queries that discard vast amounts of data, degrading synthesis efficiency. We propose a time-truncation harness that enforces a temporal cut-off at every turn, enabling TIR-style sampling from historical events, reducing temporal leakage and reliance of rejection sampling or unsolved queries, increasing the sampling efficiency. We further build a large-scale corpus and a process-based metric and show that our harness naturally induces a broader temporal breadth of search and raises the proportion of high-quality data, further increasing the efficiency and reducing the reliance on complex rubrics. Distillation experiments show that students trained on harness-intervened data achieve the best performance, demonstrating harness-assisted model evolving that turns higher quality temporal search and reasoning data into a parametric advancement of the students.
\end{abstract}

\begin{figure}[htbp]
    \centering
    \vspace{-16px}
    \makebox[\linewidth][c]{\includegraphics[width=\textwidth]{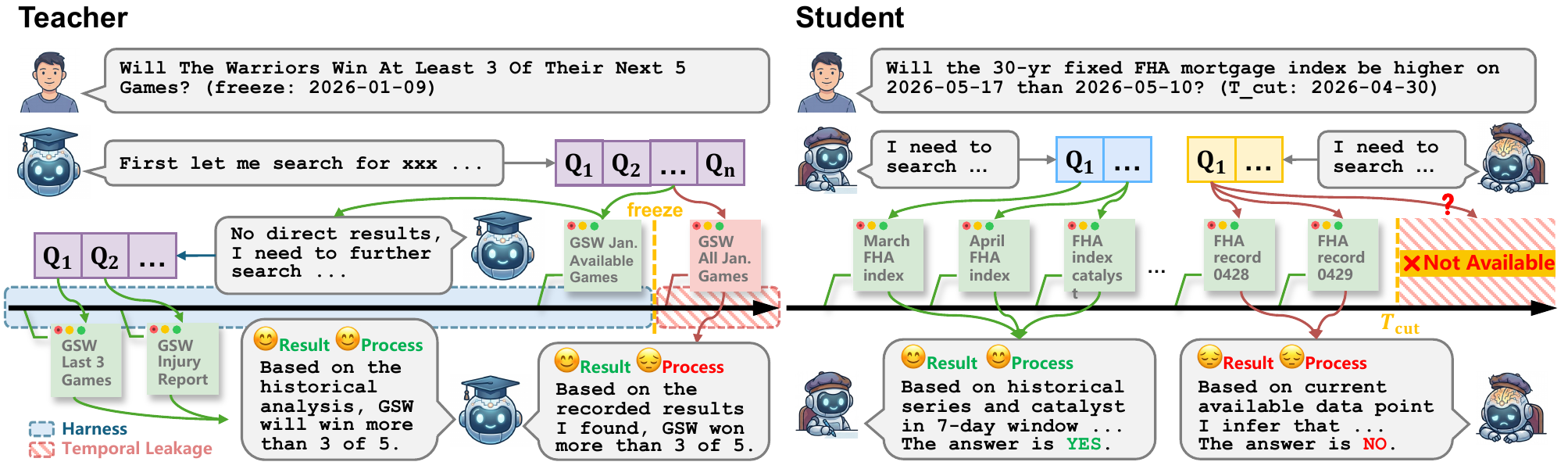}}
    \caption{\textbf{Case Study of Harness-Intervened Trajectory Synthesis and Model Evolution}. The Time-Truncation Harness Dynamically Restricts the Search Timeline (Freeze) throughout the Teacher's TIR Process.
    Blocking Future Information via the Harness Mitigates Temporal Leakage and Raises Sampling Efficiency.
    Forced to Bypass Retrieval Shortcuts, the Teacher Generates Higher-quality Temporal Search and Reasoning Trajectories. Distilling these Trajectories Converts the External Harness into the Student's Parametric Capabilities.}
    \label{fig:sell}
\end{figure}

\section{Introduction}

Predicting future events carries significant social implications across domains such as finance \citep{finance}, public health \citep{publichealth}, and politics \citep{politic}. Accurate forecasting requires synthesizing complex, dynamic information to infer unknown outcomes. Currently, state-of-the-art approaches rely on LLMs augmented with harness engineering \citep{miroflow, h2o, milkyway}. These methods primarily leverage test-time scaling, enhancing predictive accuracy by enriching the context and scaling the number of interaction branches and turns. However, such predictive capability is never internalized into the LLM's parameters. Once detached from the external harness, performance degrades significantly due to the absence of multi-turn interaction constraints and context management.

To internalize these capabilities, recent research emphasizes \textbf{Tool-Integrated Reasoning (TIR)} \citep{tir1,tir2}. Utilizing standard ReAct-style frameworks \citep{react}, models are trained to interact with environments, leverage tools for observations, and deduce conclusions. Frontier LLMs acquire general agentic capabilities through mixed-corpus pre-training and post-training \citep{glm5, kimik2.5, longcatflashthinking2601}, while smaller models typically achieve them via supervised fine-tuning (SFT) on teacher-generated trajectories and outcome-based reinforcement learning (RL) \citep{searchr1, deepdive, mirothinkerv1.7h1}. These paradigms equip models with \textbf{Deep Search} capability, defined as autonomous multi-hop retrieval over extensive information to locate static facts \citep{browsecomp, browsecompzh}. However, deep search remains insufficient for forecasting, which instead requires \textbf{Temporal Search and Reasoning}: analyzing historical trends, capturing dynamic temporal shifts, and inferring future outcomes. When handling prediction tasks, current methods primarily rely on basic deep search capabilities and generalization over forecasting queries, lacking mechanisms that explicitly internalize temporal search and reasoning ability.

Recent studies attempt to train LLMs specifically for future prediction via autonomous tool calls while reasoning. \textbf{However, acquiring dynamic training corpora is insufficient due to the time-dependent nature of forecasting}. Utilizing historical events as queries introduces temporal leakage, as models can directly search for the known outcome, degrading the task into basic information retrieval. Previous works provide static observations for historical queries to prevent leakage \citep{outcome,thinkingmatchines}, but this freezes the information-gathering process and restricts training to learning predictions from fixed prompts alone. Newer methods adopt rejection sampling \citep{echo}, but they rely on complex rubric definitions and discard a large proportion of the generated data, resulting in inefficient data synthesis. Novel works deploying real-time queries in real-world environments \citep{futureworld, echo}, which avoid temporal leakage but introduce critical challenges for RL, including severe reward sparsity since models receive only outcome-based supervision without step-wise guidance, and prohibitive latency in the training loop as the system must wait for real-world events to resolve. 

To address these issues, we propose a time-truncation harness that dynamically enforces a temporal cutoff at each turn of the agent's search, thereby restricting the accessible timeline to the period preceding the historical event during trajectory synthesis. This mechanism directly samples trajectories with a substantially higher leak-free ratio from historical queries, largely obviating the need for rejection sampling or query selection and thereby improving data-synthesis efficiency. To evaluate this approach, we collect large-scale historical forecasting queries and analyze the underlying reasoning process using a novel process-based metric that quantifies the temporal breadth of the search. Empirical results demonstrate that trajectories produced under the harness exhibit a higher metric score, indicating that the constrained teacher model naturally expands its historical information retrieval horizon under limited observations to improve search quality and yield a higher proportion of high-quality data. Finally, distilling these trajectories into student models reveals that harness-intervened data serves as better distillation inputs since the resulting students achieve higher benchmark and metric scores. This ultimately demonstrates a successful paradigm of model evolution that repurposes structured temporal search patterns into parametric advancements.
In summary, our contributions are three-fold:

\begin{itemize}
    \item We propose \textbf{a novel time-truncation harness} for TIR-style trajectory synthesis rather than static data construction, substantially reducing the need for leakage-based rejection sampling and for waiting on fresh unresolved queries, thereby improving sampling efficiency.

    \item We construct \textbf{a large-scale future prediction corpus} and introduce \textbf{a process-based metric} to quantify the temporal breadth of the search. We demonstrate that our harness naturally enhances the efficiency of sampling high-quality temporal search and reasoning data and reduces the reliance on process-evaluation rubrics.

    \item We validate the \textbf{harness-intervened model evolving} through distillation experiments on student models, showing that trajectories collected under the harness yield better performance and higher process-based metric among the compared student models on two benchmarks.
\end{itemize}

\section{Related Works}

\subsection{Agent Frameworks for Future Prediction}
Most current future-prediction systems are built as agent harnesses that orchestrate frontier LLMs, pushing capabilities to their limits. General-purpose harnesses such as Claude Code \citep{claudecode}, Codex \citep{codex}, and OpenClaw \citep{openclaw} supply basic tool-use, search, reasoning, and memory functions, readily applicable to autonomous forecasting. Others focus on multi-round deep search: OpenAI Deep-Research \citep{openaideepresearch} and Gemini Deep-Research \citep{geminideepresearch} enable planned comprehensive investigations and report generation; Flash-Searcher \citep{flashsearcher} uses a DAG-based Plan-and-Execute architecture to enhance task decomposition; MiroFlow \citep{miroflow} and H2O AI Super Agent \citep{h2o} combine sub-agent structures with heavy inference on top-tier LLMs. Recent work introduces prediction-specific harnesses such as Milkyway \citep{milkyway}, which performs multi-day iterative forecasting to refine performance. All these systems improve accuracy via test-time scaling without modifying the underlying LLM; none of them addresses tuning LLMs specifically for deep search or future prediction.

\subsection{Training LLMs for Future Prediction}
A second line of works focus on training LLMs to autonomously drive forecasting. Current frontier LLMs such as GLM-5 \citep{glm5}, Kimi-K2.5 \citep{kimik2.5}, and LongCat-Flash-Thinking-2601 \citep{longcatflashthinking2601} already support TIR to execute general multi-turn agentic tasks. For small-scale models, prior work primarily focuses on training deep search capabilities. Search-r1 \citep{searchr1} and DeepDive \citep{deepdive} utilize agentic RL to roll out in environments, computing rewards based mainly on outcomes. MiroThinker v1.7/H1 \citep{mirothinkerv1.7h1} introduces a verifier-based inference framework and performs SFT on models. However, these methods mainly target general multi-hop factual questions and overlook the specific scenario of future prediction. 

To address this, early works \citep{approaching} introduced a decoupled architecture separating retrieval and reasoning, establishing a basic forecasting system. Subsequent studies \citep{outcome,thinkingmatchines} built on this paradigm by leveraging outcomes as supervision signals to train LLMs. However, these approaches are fundamentally based on static data construction \citep{prophet,llmasprophet,futureaslabel}, where each query observation is fixed, and therefore cannot support tool-integrated reasoning during training. To solve this problem, FutureWorld \citep{futureworld} and Echo \citep{echo} use agents to continuously collect unsolved prediction queries and compute rewards once the results are available. However, the online RL stage inevitably suffers from reward sparsity and high latency. Echo \citep{echo} additionally searches for domain-specific rubrics automatically to select trajectories, but such customized rubrics incur substantial data-selection costs.

\section{Preliminaries}
\label{sec:prelim}

\paragraph{Tool-Integrated Reasoning}
Given a fixed system prompt $S_0$ and a fixed query $P$, a TIR-style trajectory is:
\begin{equation}
\tau_{tir} := (S_0,P,a_1,o_1,a_2,o_2,\dots,a_T), \
              a_{i(i<T)}=\{r_i,t_i\}, \ o_{i(i<T)}=\mathcal{E}(t_i), \ a_T=\{r_T,S\}
\label{eq:tir_traj}
\end{equation}
where $a_i$ represents the $i$-th action produced by the model: for intermediate steps $i < T$, each action $a_i$ is composed of reasoning content $r_i$ and a tool call $t_i$. The corresponding $o_i$ is the observation returned after executing $t_i$ from environment $\mathcal{E}$. $a_T$ represents the terminal action, where $S$ is the final answer generated by the model without further tool invocation.

\paragraph{Tools for Search Agent} The agent is equipped with a unified tool environment consisting of two core actions: $t_i \in \{\texttt{Search,Browse with Goal}\}$. (1) \textbf{Search}, which allows the agent to issue a batch of up to five queries per invocation, and (2) \textbf{Browse with Goal}, which allows the agent to navigate to a given webpage under a specified goal and returns a goal-conditioned summary of that page. Detailed tool schemas are provided in Appendix~\ref{appendix:tools}.

\paragraph{Temporal Leakage} For forecasting, sampling from historical events whose outcomes are already known raises the risk of temporal leakage: the environment may return documents that postdate the event and reveal the answer. For document $d$, let $\rho(d)$ be the publication time of $d$, temporal leakage can be defined as:
\begin{equation}
    \label{eq:leakage}
    \text{TempLeak}\big(o_i, T_\text{end}\big) = \mathbb{I}\big[\exists\, d \in \mathcal{E}(t_i) \ \text{ s.t. } \ \rho(d) \ge T_\text{end}\big]
\end{equation}

\paragraph{Sampling Efficiency}
A data-synthesis approach $\mathcal{A}$ should yield as many valid trajectories as possible per unit cost:
\begin{equation}
J(\mathcal{A})
=\underbrace{\mathrm{Throughput}(\mathcal{A})}_{\text{queries per round}}
\times
\underbrace{\bigl(1-\mathrm{RejectionRate}(\mathcal{A})\bigr)}_{\text{valid fraction}}
\label{eq:efficiency}
\end{equation}
where $\mathrm{Throughput}$ denotes the rate at which trajectories are generated, and $\mathrm{RejectionRate}$ denotes the fraction of trajectories discarded by rejection sampling. The complementary term $\bigl(1-\mathrm{RejectionRate}(\mathcal{A})\bigr)$
therefore corresponds to the fraction of trajectories retained as valid supervision.

\section{Methodology}
\label{sec:method}

\subsection{Limitations of Prior Works}
\label{sec:limitationsofprior}
\paragraph{Static Observation}
A common practice abandons agentic interaction and instead feeds the model a set of pre-collected observations. Under this paradigm, the trajectory reduces to:
\begin{equation}
\tau_{\text{static}} := (S_0, P, o_{all},a), \ a=\{r,S\}
\label{eq:static}
\end{equation}
Unlike the TIR-style trajectory $\tau_{tir}$ in Eq.~\eqref{eq:tir_traj}, the static formulation collapses the reasoning process into a single-step inference conditioned on a fixed observation set $o_{all}$. As a result, the model thus loses the ability to dynamically query, refine, or update information based on intermediate reasoning states. 

\paragraph{Rejection Sampling} This method draws trajectories $\tau$ from the unconstrained environment $\mathcal{E}$ and subsequently discards any trajectory that exhibits temporal leakage (Eq.~\eqref{eq:leakage}) or violate a predefined quality rubric $\mathcal{R}(\tau) = 0$. The set of accepted trajectories $\mathcal{T}_{\text{accepted}}$ is formally defined as:
\begin{equation}
\label{eq:rejectionsampling}
\mathcal{T}_{\text{accepted}} = \Big\{ \tau \ \Big| \ \text{TempLeak}\big(d, T_\text{end}\big) = 0 \ \wedge \ \mathcal{R}(\tau) = 1 \Big\}
\end{equation}
This filtering process causes the rejection rate to surge, driving $\bigl(1-\mathrm{RejectionRate}(\mathcal{A})\bigr)$ toward $0$ in Eq.~\eqref{eq:efficiency}.

\paragraph{Unresolved Queries} Let $\mathcal{Q}$ be the global query set and $T_{\text{now}}$ be the current operational timestamp. This strategy restricts the selection to the active query subset $\mathcal{Q}_{\text{active}}$:
\begin{equation}
\label{eq:unsolvedqueries}
\mathcal{Q}_{\text{active}} = \Big\{ q \in \mathcal{Q} \ \Big| \ T_{\text{end}}(q) > T_{\text{now}} \Big\}
\end{equation}
While this constraint guarantees $\text{TempLeak} = 0$ in Eq.~\eqref{eq:leakage}, it forces the pipeline to wait for real-world event outcomes before a trajectory can be used as a labeled sample, thus $\mathrm{Throughput}(\mathcal{A}) \downarrow$ in Eq.~\eqref{eq:efficiency}.

\subsection{Our Solution: Time-Truncation Harness Design}
\label{sec:harness}

Rather than removing the agentic interaction (Eq.~\eqref{eq:static}) or filtering queries and trajectories (Eq.~\eqref{eq:rejectionsampling} and \eqref{eq:unsolvedqueries}), we constrain the search environment itself so that the agent sees only information available before the event at every turn.

\paragraph{Cut-Off Date}
For a query whose event resolves at $T_\text{end}$, we set a cut-off date:
\begin{equation}
T_\text{cut} = T_\text{end} - \Delta_T ,
\label{eq:cutoff}
\end{equation}
where the margin $\Delta_T > 0$ terminates the search window strictly before the event resolves.

\begin{figure}
    \centering
    \vspace{-16px}
    \includegraphics[width=\linewidth]{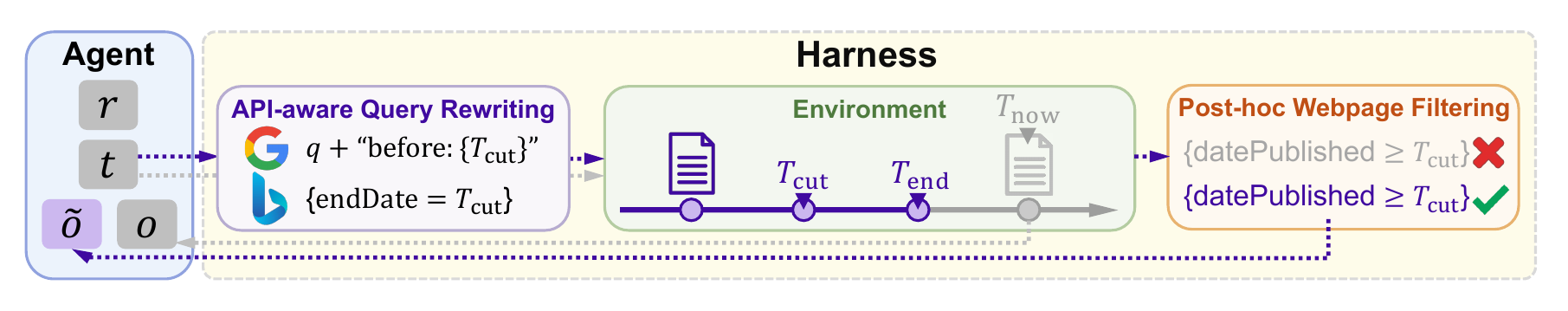}
    \caption{\textbf{Overview of the Time-Truncation Harness}. The Framework Demonstrates the Comparison Between the Standard Unconstrained Search Path (\textbf{\textcolor{linegray}{Gray Line}}) and the Proposed Harness-Intervened Path (\textbf{\textcolor{linepurple}{Purple Line}}) in One Round of Agent-Environment Interaction $r \rightarrow t\rightarrow \textcolor{linegray}{\textbf{o}} \ \text{or} \ \textcolor{linepurple}{\tilde{\textbf{o}}}$.}
    \label{fig:harness}
\end{figure}

\paragraph{Time-Constrained Environment}
We replace $\mathcal{E}$ with a truncated environment $\mathcal{E}_{T_\text{cut}}$ that
returns only documents published no later than $T_\text{cut}$:
\begin{equation}
\tilde{o}_i = \mathcal{E}_{T_\text{cut}}(\tilde{t}_i)
= \bigl\{\, d \in \mathcal{E}(\tilde{t}_i) : \rho(d) \le T_\text{cut} \,\bigr\}
\label{eq:trunc-env}
\end{equation}
As shown in Figure~\ref{fig:harness}, $\mathcal{E}_{T_\text{cut}}$ is realized in two complementary ways:
\textbf{API-aware Query Rewriting} pushes the bound into each search backend through its native interface 
so results are constrained at the source; and a
\textbf{Post-hoc Webpage Filter} drops any returned item whose publication time (extracted from the \texttt{datePublished} field) is missing, unparsable, or later than $T_\text{cut}$, serving as a more solid guarantee. Compared with Eq.~\eqref{eq:leakage}, this reduces the temporal leakage.

\paragraph{Aligned System Prompt}
To align the agent with the truncated environment, we design system prompt to help the model better adapt to the environment on \(T_\text{cut}\):
\begin{equation}
\tilde{S_0} = S_0 \oplus \mathcal{I}(T_\text{cut}),
\end{equation}
where $\mathcal{I}(T_\text{cut})$ denotes the injected instructions that keep the agent's stated temporal context consistent with the harness setting.
The full prompts are provided in Appendix~\ref{appendix:system prompt}.

\section{Data Preparation}
\label{sec:data}

\subsection{Query Collection}
\label{query}
We construct \textbf{12{,}378} queries \footnote{\url{https://huggingface.co/datasets/wxcai/manifold_newest_multi_domains_260318}} from \textbf{Manifold Market} (by its official API\footnote{\url{https://docs.manifold.markets/api}}), a public forecasting platform providing crowd-sourced probabilities and resolved ground truths. 
We utilize \textbf{Google Serper API}\footnote{\url{https://serper.dev}}
for all rollouts.
We filter the evaluation markets based on four core criteria: (1) We maintain \textbf{Verifiability} by retaining only fully resolved markets to guarantee definitive outcomes. (2) To ensure \textbf{Objectivity}, we exclude subjective queries and opinion polls, focusing exclusively on events determined by external reality. (3) We emphasize \textbf{Diversity}, spanning 30 real-world domains (e.g. technology, politics, economics) across 4 types of questions (binary, multiple-choice, date and numeric). (4) We \textbf{Eliminate Pre-training Contamination} by selecting markets resolved between June 2025 and March 2026, a window prior to our benchmark evaluation and beyond the student model's knowledge cutoff in following distillation experiment. The composition is summarized in Figure~\ref{fig:corpuscomposition} and more details of construction are provided in Appendix~\ref{appendix:data}.

\subsection{Corpus Construction}
\label{corpus}
Following the query collection process, we leverage \textbf{Kimi-K2.5} \citep{kimik2.5} as our advanced teacher model to synthesize trajectories. To investigate the explicit behavioral and distributional effects of temporal constraints, we establish two comparative groups. Additional configurations are detailed in Appendix~\ref{appendix:teacher_config}.
\begin{itemize}
    \item \textit{W/o-Harness Group:} The teacher model generates trajectories under standard unconstrained environment settings ($\mathcal{E}$), where it can freely retrieve documents.
    \item \textit{W/-Harness Group:} The teacher model interacts with the time-truncation harness environment ($\mathcal{E}_{T_{\text{cut}}}$). To absorb the delay between an event's occurrence and its market settlement, we enforce a pre-event temporal margin by setting $\Delta_T = 3\text{ days}$
\end{itemize}

\begin{figure}
    \centering
    \vspace{-16px}
    \includegraphics[width=0.9\linewidth]{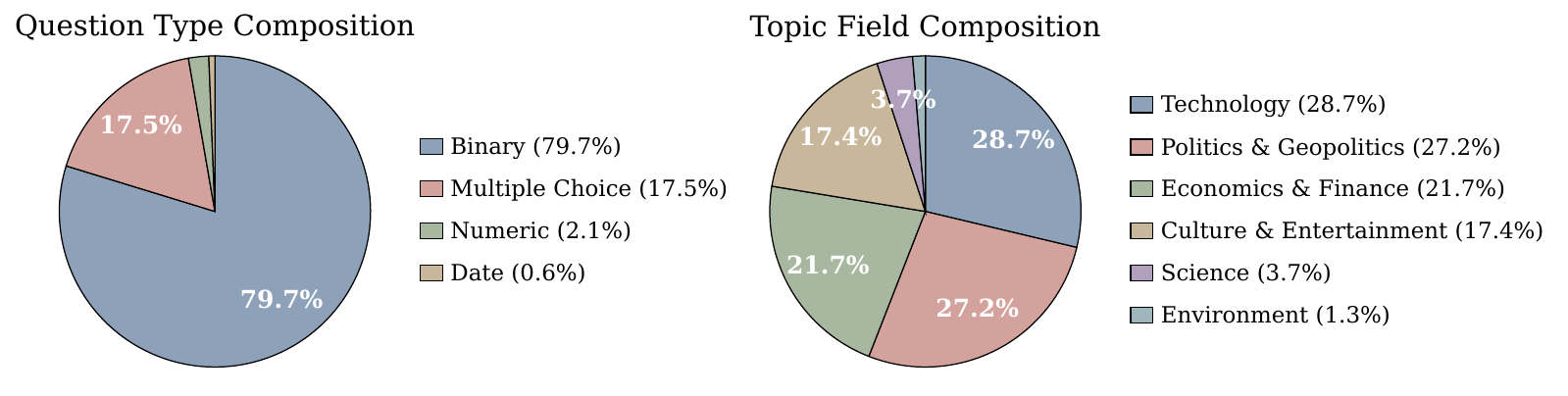}
    \caption{Composition of the Manifold Corpus by Question Type (Left) and Topic Field (Right).}
    \label{fig:corpuscomposition}
\end{figure}

\section{Experiments}
\label{sec:experiments}

\subsection{Temporal Leakage and Efficiency Analysis}
Under the harness, a time-constrained trajectory is:
\begin{equation}
\label{eq:harness_tir}
\tilde{\tau}_{tir}
:=
(\tilde{S_0}, P, \tilde{a_1}, \tilde{o}_1, \tilde{a_2}, \tilde{o_2}, \dots, \tilde{a}_i), \
\tilde{a}_{i(i<T)} = \{\tilde{r}_t, \tilde{t}_i\}, \ 
\tilde{o}_i = \mathcal{E}_{T_\text{cut}}(\tilde{t}_i).
\end{equation}
This construction alleviates the following three issues:
\begin{itemize}
\item \textbf{Dynamics.} The sequential $\tilde{o}_i$ preserve multi-turn TIR style (Eq.~\eqref{eq:tir_traj}) in contrast to static observation settings (Eq.~\eqref{eq:static}).
\item \textbf{Low-Rejection.} TempLeak in Eq.~\eqref{eq:leakage} is relieved, reducing rejection sampling (Eq.~\eqref{eq:rejectionsampling}) and ensuring more trajectories are valid. Thus $\bigl(1-\mathrm{RejectionRate(\mathcal{A})}\bigr) \uparrow$ in Eq.\eqref{eq:efficiency}.
\item \textbf{High-Throughput.} All historical queries $\mathcal{Q}$ (Eq.~\eqref{eq:unsolvedqueries}) become immediately usable as supervision under the harness, decoupling data collection from event resolution $\mathcal{Q}_{\text{active}}$ and enabling $\mathrm{Throughput}(\mathcal{A})\uparrow$ in Eq.\eqref{eq:efficiency}.
\end{itemize}

\begin{table}[htbp]
\centering
\vspace{-10px}
\caption{Leakage Statistics on Randomly Sampled Subset.}
\vspace{4px}
\label{tab:leakage_efficiency_analysis}
\resizebox{0.6\linewidth}{!}{
\begin{tabular}{lcccc}
\toprule
\textbf{Dataset} & \textbf{Sample Number} & 
\textbf{Date-Leak} & \textbf{Content-Leak} & \textbf{Valid(\%)} \\
\midrule
w/ harness & 200 & \textbf{0} & \textbf{23} & \textbf{88.5\%}  \\
w/o harness & 200 & 181 & 101 & 7.0\%  \\
\bottomrule
\end{tabular}
}
\end{table}

To validate the claim, we randomly select 200 samples from each group and design a two-step verification procedure. For each round within a trajectory, we first inspect the \texttt{datePublished} field of each search result. If $\texttt{datePublished} \ge T_{\text{cut}}$, the instance is directly flagged as leaked. For entries with $\texttt{datePublished} < T_{\text{cut}}$, we conduct an additional content audit with Claude-Opus-4.8 \citep{claudeopus4.8} (prompt provided in Appendix~\ref{appendix:verify prompt}) to determine whether the content contains description referencing events after $T_{\text{cut}}$. Results are summarized in Table~\ref{tab:leakage_efficiency_analysis}. These results show that the harness eliminate \texttt{datePublished}-level leakage and substantially reduces the content-level leakage, thereby increasing the amount of valid training data synthesized from historical queries.

\subsection{Corpus Statistical Analysis}
\label{sec:corpusanalysis}

When the time-truncation harness is imposed, the transition dynamics of the environment shift from $P(o_t \mid a_t, \mathcal{E})$ to $P(\tilde{o}_t \mid a_t, \mathcal{E}_{T_{\text{cut}}})$, where any document $d$ with a publication timestamp $\rho(d) > T_{\text{cut}}$ is strictly removed. Consequently, the conditional trajectory distribution transforms into:
\begin{equation}
    P_\theta(\tilde{\tau}) = \prod_{t=1}^{T} P_\theta(\tilde{a}_t \mid \tilde{S}_0, P, \tilde{a}_{<t}, \tilde{o}_{<t}) \cdot P(\tilde{o}_t \mid \tilde{a}_t, \mathcal{E}_{T_{\text{cut}}})
\end{equation}
To maintain optimization under the change of tool results in each turn, the policy $P_\theta$ is forced to shift. 
To investigate how reasoning and search queries adapt to changes in observations, we first employ a basic outcome-based metric, \textbf{Accuracy}, and a basic process-based metric, \textbf{Tool-Call Count}. However, neither directly evidences how the search process itself changes. We therefore design a dedicated process-based metric, the \textbf{Query Date-Span}.
Let $\mathcal{Q}(\tau)$ be the set of search queries in trajectory $\tau$,
and let $\mathrm{Dates}(q)$ return the calendar dates parsed from query $q$
(e.g. \texttt{"Jan.\ 1 2026"} or \texttt{"Jan.\ 2026"}, details are given in Appendix~\ref{appendix:querydatespan}). The Query Date-Span (in days) is:
\begin{equation}
  \mathcal{M}(\tau)=
    \max\mathcal{D}(\tau)-\min\mathcal{D}(\tau), \  \mathcal{D}(\tau)=\bigcup_{q\in\mathcal{Q}(\tau)}\mathrm{Dates}(q)
\end{equation}

\begin{figure}
    \centering
    \vspace{-16px}
    \includegraphics[width=\linewidth]{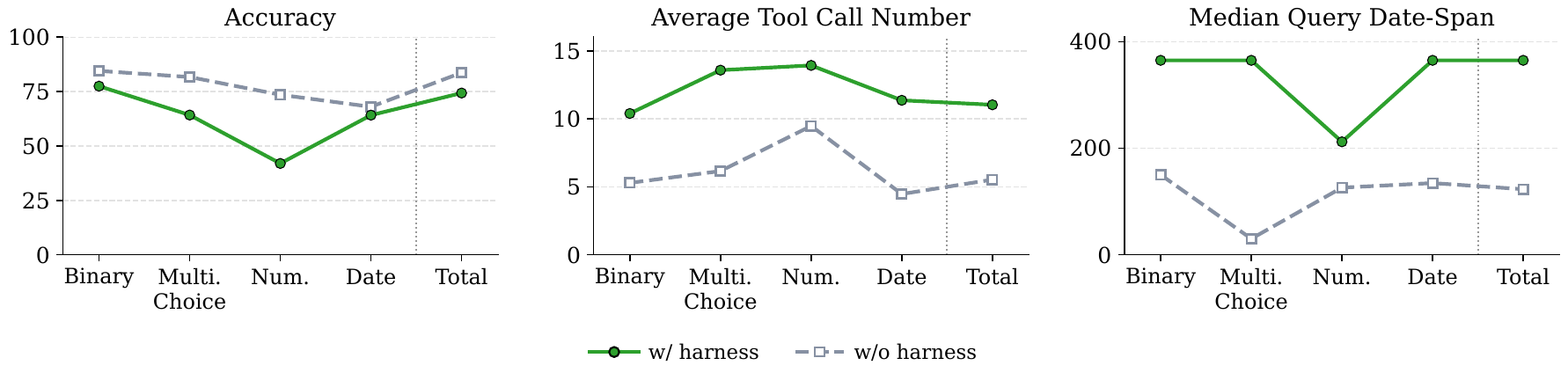}
    \caption{Comparison of Accuracy, Average Tool Call Number and Median Query Date-Span for Kimi-K2.5 Synthesized Corpora with and without Time-Truncation Harness.}
    \label{fig:corpus}
\end{figure}

To understand the precise impact of the time-truncation harness on the generated trajectories, we conduct an extensive empirical analysis across the two generated corpora. The results are illustrated in Figure \ref{fig:corpus}.

\paragraph{Accuracy Decrease}
We first examine how the corpus changes with and without the harness from an outcome perspective. The final answer accuracy exhibits an expected reduction when the harness is applied (\textit{w/ time-truncation}) compared to the unconstrained baseline (\textit{w/o time-truncation}). Across all question types, the accuracy degradation ranges approximately from 5\% to 25\%. The observed accuracy drop implies that temporal leakage affords the model shortcut opportunities on a portion of samples, whereas the harnessed setting renders such shortcuts unavailable and thus proves fundamentally more challenging.

\paragraph{Expansion of Tool Call Numbers}
Confronted with this increased difficulty and the scarcity of direct shortcuts, the model is forced to gather information more exhaustively. Consequently, we observe a substantial increase in the number of search tool calls. Unable to retrieve direct outcomes in a single step, the agent must iteratively query the search engine. This expansion in the number of search rounds in turn drives a shift in the model's underlying querying strategy.

\paragraph{Expansion of Temporal Search and Reasoning}
Built upon these extended search interactions, the trajectories produced under the harness show a massive increase in the median Query Date-Span, exceeding 300 days on average across nearly all categories, whereas the unconstrained group stays below 150 days. These behavioral divergences suggest that when direct answers are largely blocked, the teacher model naturally expands its temporal retrieval horizon. It actively searches for broader historical context, baseline data, and longitudinal trends spanning months prior to the event to perform rational deduction. Therefore, although the harness lowers absolute accuracy, it raises the proportion of high-quality temporal search and reasoning trajectories within the corpus.

\subsection{Distillation Experiments}

Given the trajectory-level differences induced by the harness, a natural question is whether these behavioural shifts translate into substantive training gains. Therefore, we conduct distillation experiments to investigate how these variations impact the internalization of capabilities within student models.

\subsubsection{Experimental Setup}
\paragraph{Model Selection} We employ two standard open-source thinking models as our student models: Qwen3-8B \citep{qwen3} and Qwen3-32B \citep{qwen3}, both trained via \textbf{full-parameter supervised fine-tuning (SFT)}. Given that the Qwen3 backbone exhibits limited native multi-turn tool-calling capability, fine-tuning solely on the 10K-scale forecasting corpus risks format degradation and execution failures during TIR rollouts. To maintain robust behaviors, we introduce \textbf{a mixed-corpus SFT strategy} that incorporates standard general deep-search trajectories.

\paragraph{Data Mixture} We first curate a baseline group by filtering 50,000 multi-hop deep-search trajectories with context lengths strictly within 65K tokens. For the forecasting corpus, we retain trajectories whose sequence length is below 65K tokens. To equalize data volume across groups, the remaining slots are filled by randomly sampling from the 50,000 general deep-search trajectories until each mixture contains exactly 50,000 instances. More information about trajectory format processing, parameter settings and loss is detailed in Appendix \ref{appendix:sft}.
All these steps result in three groups of SFT data:
\begin{enumerate}
    \item \textit{DeepSearch-SFT}: trained on the 50,000 general multi-hop deep-search trajectories, serving as our baseline.
    \item \textit{Mix-SFT (w/o harness)}: fine-tuned on 11,868 valid unconstrained forecasting trajectories and 38,132 randomly sampled deep-search instances (total 50,000).
    \item \textit{Mix-SFT (w/ harness)}: leveraging the identical mixture structure but utilizing 11,261 valid harness-intervened trajectories padded with 38,739 randomly sampled deep-search instances (total 50,000).
\end{enumerate}

\paragraph{Benchmarks and Metrics}
We survey related benchmarks (summarized in Appendix~\ref{appendix:benchmarks}) and select the two most suitable ones. ForecastBench \citep{forecastbench} is a monthly-updated dynamic benchmark that spans 8 real-world domains. We use five OOD domains to ensure adequate data and avoid contamination, collecting the questions that resolve in May 2026. The metric used in ForecastBench is \textbf{Brier Score}. FutureX \citep{futurex} is a weekly-updated dynamic benchmark that contains 4 levels based on the difficulty (L1,L2 for single/multiple choice and L3,L4 for ranking and numeric forecasting). For FutureX, we adopt the \textbf{Score} defined in the original paper. More details of these two benchmarks and the following metrics are summarized in Appendix \ref{appendix:benchmarksettings}.

    

\subsubsection{Main Results on ForecastBench}

\begin{table}[htbp]
\centering
\caption{\textbf{Probabilistic Forecasting Performance of Distilled Student Models on ForecastBench (May) across 5 Different Domains}. Lower Brier Score Indicates Better Performance. Results are Averaged over 4 Independent Runs (Mean $\pm$ Std). The Best Results are Highlighted in \textcolor{red}{\textbf{RED}}.}
\label{tab:forecastbench_results}
\vspace{4px}
\setlength{\tabcolsep}{3.5pt}
\resizebox{\textwidth}{!}{
\begin{tabular}{lcccccc}
\hline
\multirow{2}{*}{\textbf{Model \& Setting}} & \multicolumn{6}{c}{\textbf{ForecastBench (May)}} \\
 & \textbf{Total (246)} & \textbf{ACLED (50)} & \textbf{DBNOMICS (49)} & \textbf{FRED (48)} & \textbf{Wikipedia (49)} & \textbf{yfinance (50)} \\ \hline
\textit{\textbf{Qwen3-8B Backbone}} & & & & & & \\
\quad DeepSearch-SFT & 0.4344 $\pm$ 0.0125 & 0.5113 $\pm$ 0.0184 & 0.6099 $\pm$ 0.0211 & 0.3896 $\pm$ 0.0105 & 0.2367 $\pm$ 0.0094 & 0.4215 $\pm$ 0.0162 \\
\quad Mix-SFT (w/o harness) & 0.3167 $\pm$ 0.0098 & 0.2893 $\pm$ 0.0142 & 0.4676 $\pm$ 0.0195 & 0.3360 $\pm$ 0.0118 & 0.1490 $\pm$ 0.0076 & 0.3421 $\pm$ 0.0133 \\
\quad Mix-SFT (w/ harness) & \textcolor{red}{\textbf{0.2550}} $\pm$ 0.0064 & \textcolor{red}{\textbf{0.2334}} $\pm$ 0.0091 & \textcolor{red}{\textbf{0.3576}} $\pm$ 0.0124 & \textcolor{red}{\textbf{0.3047}} $\pm$ 0.0083 & \textcolor{red}{\textbf{0.0965}} $\pm$ 0.0041 & \textcolor{red}{\textbf{0.2838}} $\pm$ 0.0097 \\ \hline
\textit{\textbf{Qwen3-32B Backbone}} & & & & & & \\
\quad DeepSearch-SFT & 0.4365 $\pm$ 0.0112 & 0.3473 $\pm$ 0.0156 & 0.6988 $\pm$ 0.0245 & 0.4088 $\pm$ 0.0139 & 0.2014 $\pm$ 0.0082 & 0.5254 $\pm$ 0.0191 \\
\quad Mix-SFT (w/o harness) & 0.2563 $\pm$ 0.0071 & 0.2775 $\pm$ 0.0104 & 0.3808 $\pm$ 0.0152 & 0.2669 $\pm$ 0.0092 & 0.0535 $\pm$ 0.0033 & 0.3019 $\pm$ 0.0115 \\
\quad Mix-SFT (w/ harness) & \textcolor{red}{\textbf{0.1998}} $\pm$ 0.0043 & \textcolor{red}{\textbf{0.1548}} $\pm$ 0.0067 & \textcolor{red}{\textbf{0.3300}} $\pm$ 0.0098 & \textcolor{red}{\textbf{0.2198}} $\pm$ 0.0054 & \textcolor{red}{\textbf{0.0364}} $\pm$ 0.0019 & \textcolor{red}{\textbf{0.2580}} $\pm$ 0.0078 \\
\hline
\end{tabular}

}
\vspace{4px}
\begin{tablenotes}
    \footnotesize
    \item \textit{Note: The original model fails to perform multi-round tool calling and tends to output a uniform probability of 0.5, rendering its Brier Score meaningless.}
\end{tablenotes}
\end{table}

\paragraph{Advantage of Harness-Intervened Data}
As shown in Table~\ref{tab:forecastbench_results}, students trained with harness-generated trajectories consistently achieve the lowest Brier Scores. Compared with \textit{Mix-SFT (w/o harness)}, \textit{Mix-SFT (w/ harness)} improves overall performance by \textbf{19.48\%} and \textbf{22.04\%} under 8B and 32B backbones, respectively. These gains indicate that harness-intervened trajectories help students acquire genuine temporal reasoning and forecasting behaviour by reducing temporal leakage and improving search quality, rather than by exploiting post-event information. Overall, the time-truncation harness offers an effective route to distilling forecasting ability into smaller models.

\paragraph{Mixing Alone is Not the Key}
Comparing the \textit{Mix-SFT} variants with \textit{DeepSearch-SFT} further confirms the importance of the time-truncation harness. Although \textit{Mix-SFT (w/o harness)} outperforms \textit{DeepSearch-SFT} by better aligning the training distribution with forecasting queries, the further gains of \textit{Mix-SFT (w/ harness)} show that data mixing alone is insufficient. The time-truncation harness remains essential for improving future prediction performance.

\subsubsection{Main Results on FutureX}
\label{sec:mainresults}

\begin{table}[htbp]
\centering
\caption{\textbf{Predictive Score of Distilled Student Models on FutureX across L1--L4}. Higher Score Indicates Better Performance. Results are Averaged over 4 Independent Runs (Mean $\pm$ Std). The Best Results are Highlighted in \textcolor{red}{\textbf{RED}}.}
\label{tab:futurex_results}

\vspace{4px}

\setlength{\tabcolsep}{3.5pt}
\resizebox{\textwidth}{!}{%

\begin{tabular}{lcccc|cccc}
\hline
\multirow{2}{*}{\textbf{Model \& Setting}} & \multicolumn{4}{c|}{\textbf{FutureX (April)}} & \multicolumn{4}{c}{\textbf{FutureX (May)}} \\
 & \textbf{L1 (41)} & \textbf{L2 (49)} & \textbf{L3 (24)} & \textbf{L4 (49)} & \textbf{L1 (35)} & \textbf{L2 (60)} & \textbf{L3 (44)} & \textbf{L4 (57)} \\ \hline
\textit{\textbf{Qwen3-8B Backbone}} & & & & & & & & \\
\; Origin & 71.34 $\pm$ 4.17 & 46.20 $\pm$ 2.40 & 18.17 $\pm$ 5.03 & 7.34 $\pm$ 0.94 & 66.43 $\pm$ 5.89 & 46.63 $\pm$ 5.11 & 14.03 $\pm$ 3.20 & 5.96 $\pm$ 2.84 \\
\; DeepSearch-SFT & 64.02 $\pm$ 5.03 & 47.76 $\pm$ 3.03 & 20.74 $\pm$ 3.39 & 13.88 $\pm$ 2.12 & 63.57 $\pm$ 4.88 & 54.12 $\pm$ 3.00 & 14.11 $\pm$ 1.09 & 17.76 $\pm$ 3.36 \\
\; Mix-SFT (w/o harness) & \textcolor{red}{\textbf{70.12}} $\pm$ 3.66 & \textcolor{red}{\textbf{51.22}} $\pm$ 4.89 & 19.36 $\pm$ 2.17 & 18.28 $\pm$ 2.67 & \textcolor{red}{\textbf{72.14}} $\pm$ 2.74 & \textcolor{red}{\textbf{56.83}} $\pm$ 5.64 & 18.39 $\pm$ 0.38 & 18.74 $\pm$ 1.35 \\
\; Mix-SFT (w/ harness) & 67.07 $\pm$ 6.45 & 48.09 $\pm$ 3.19 & \textcolor{red}{\textbf{21.14}} $\pm$ 7.28 & \textcolor{red}{\textbf{21.52}} $\pm$ 2.55 & 70.71 $\pm$ 6.75 & 53.78 $\pm$ 3.83 & \textcolor{red}{\textbf{19.76}} $\pm$ 1.21 & \textcolor{red}{\textbf{20.65}} $\pm$ 1.29 \\ \hline
\textit{\textbf{Qwen3-32B Backbone}} & & & & & & & & \\
\; Origin & 62.20 $\pm$ 4.22 & 48.84 $\pm$ 4.88 & 17.64 $\pm$ 5.65 & 10.41 $\pm$ 1.51 & 66.43 $\pm$ 9.72 & 58.95 $\pm$ 3.78 & 13.25 $\pm$ 3.35 & 11.94 $\pm$ 2.48 \\
\; DeepSearch-SFT & \textcolor{red}{\textbf{73.17}} $\pm$ 6.30 & 52.60 $\pm$ 7.40 & 24.21 $\pm$ 1.64 & 19.33 $\pm$ 4.18 & \textcolor{red}{\textbf{80.71}} $\pm$ 4.29 & \textcolor{red}{\textbf{75.47}} $\pm$ 1.09 & 16.04 $\pm$ 2.50 & 27.41 $\pm$ 2.78 \\
\; Mix-SFT (w/o harness) & \textcolor{red}{\textbf{73.17}} $\pm$ 0.00 & \textcolor{red}{\textbf{55.93}} $\pm$ 5.37 & 21.80 $\pm$ 1.15 & 21.17 $\pm$ 4.96 & 75.71 $\pm$ 1.65 & 81.09 $\pm$ 2.89 & 19.21 $\pm$ 1.16 & 22.77 $\pm$ 3.43 \\
\; Mix-SFT (w/ harness) & 70.85 $\pm$ 5.27 & 55.12 $\pm$ 4.10 & \textcolor{red}{\textbf{27.27}} $\pm$ 6.32 & \textcolor{red}{\textbf{23.37}} $\pm$ 2.39 & 77.14 $\pm$ 2.02 & 74.41 $\pm$ 2.97 & \textcolor{red}{\textbf{20.58}} $\pm$ 0.49 & \textcolor{red}{\textbf{27.56}} $\pm$ 3.10 \\ \hline
\end{tabular}
}
\end{table}

\paragraph{Advantage on Difficult Tasks}
As Table~\ref{tab:futurex_results} shows, L3--L4 query results on FutureX mirror those on ForecastBench: \textit{Mix-SFT (w/ harness)} consistently outperforms all baselines across both April and May slices. With an 8B backbone, it leads baselines by \textbf{13.46\%} (April) and \textbf{8.82\%} (May); at 32B, gains reach \textbf{17.74\%} and \textbf{14.08\%}.
These results confirm that the harness-intervened data remains effective on the more difficult FutureX queries.

\paragraph{Degradation on Easy Tasks}
The performance advantage narrows on lower-difficulty levels (L1--L2), where \textit{Mix-SFT (w/ harness)} shows no distinct gain and occasionally underperforms the non-harness or \textit{DeepSearch-SFT} baselines. Since simpler tasks require little information to answer, enforcing broad temporal search via the harness introduces extraneous context noise and slightly degrades accuracy when extensive extrapolation is unnecessary.

\begin{figure}
    \centering
    \vspace{-16px}
    \includegraphics[width=\linewidth]{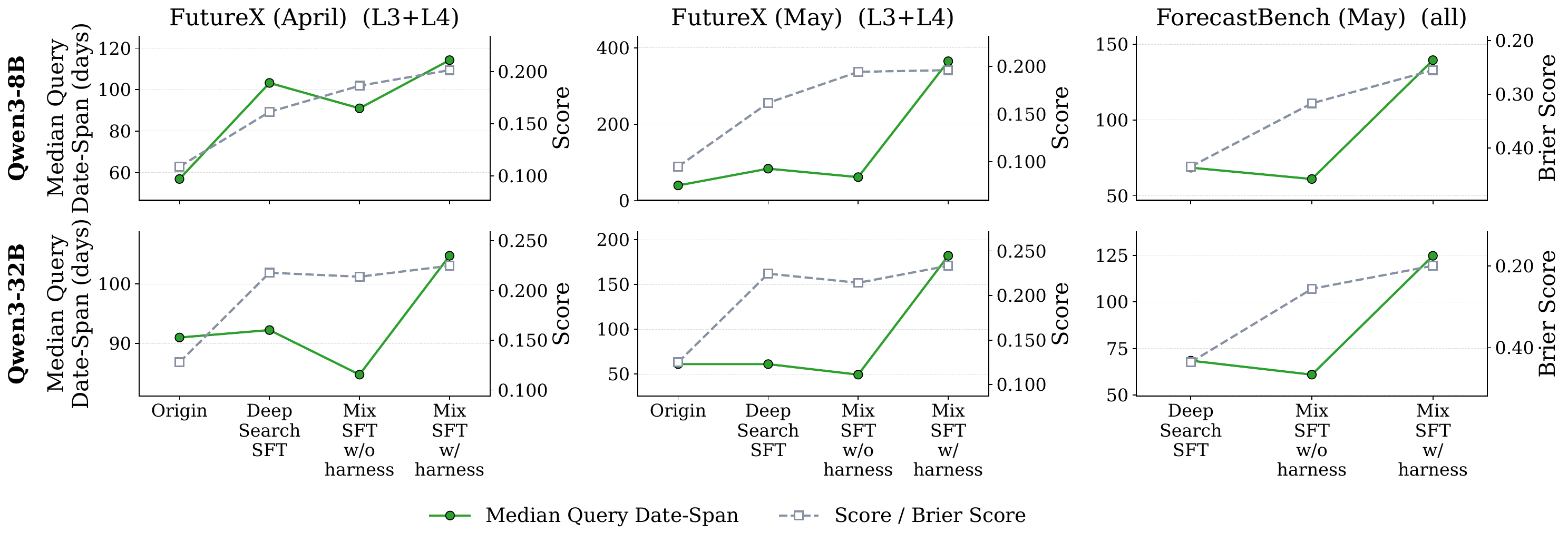}
    \caption{Correlation between Final Predictive Performance and Median Query Date-Span. We Repeat the Experiment 4 Times and Compute the Mean Value.}
    \label{fig:studentmetric}
\end{figure}

\subsubsection{Process-based Metric on Student Models}
\label{sec:studentprocess}

To further evaluate the intrinsic search behaviour of the distilled models beyond final outcomes, we analyse the median Query Date-Span of the student models using the same protocol applied to the training corpus. The results are summarized in Figure~\ref{fig:studentmetric}.

\paragraph{Teacher-Student Correlation on Temporal Search and Reasoning}
The \textit{Mix-SFT (w/ harness)} group exhibits the highest median Query Date-Span across all evaluation samples. In contrast, removing the temporal constraint or excluding temporal search and reasoning data from the training corpus yields a noticeable drop: both \textit{Mix-SFT (w/o harness)} and \textit{DeepSearch-SFT} exhibit a lower temporal span. This indicates that constraining the teacher model transfers the same behavioural characteristic to the student, which attains the broadest temporal search coverage among all groups.

\paragraph{Drawbacks of Leakage-Induced Shortcutting} The baseline \textit{DeepSearch-SFT} achieves a slightly higher median date span than the \textit{Mix-SFT (w/o harness)} variant. 
Since the general deep search paradigm inherently emphasizes multi-hop, multi-perspective retrieval trajectories, one dimension of this capability naturally manifests as an expansion across the temporal axis. 
However, when trained on unconstrained temporal search corpora, models tend to favour queries drawn from a narrow window close to the event date, which ultimately compromises the diversity and breadth of temporal search.
In summary, the temporal coverage that general deep search attains through generalization is even broader than that of models trained on leakage-prone mixed corpora.

\begin{figure}
    \centering
    \vspace{-16px}
    \includegraphics[width=\linewidth]{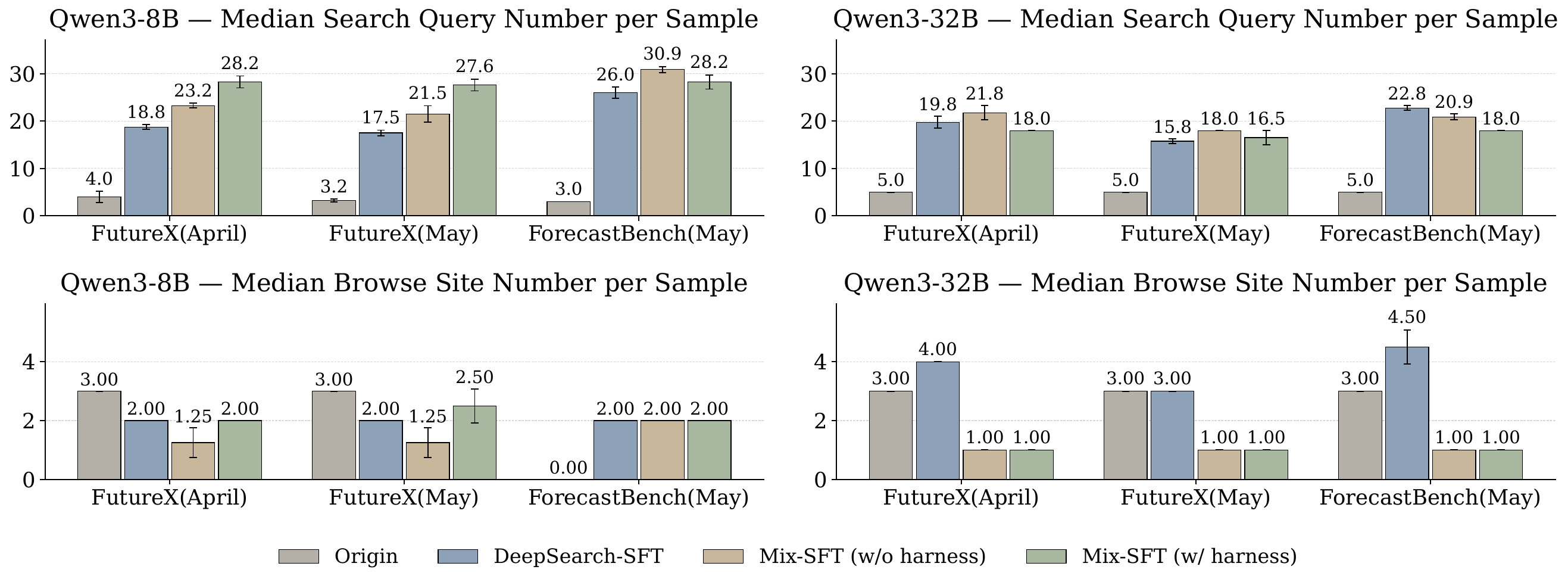}
    \caption{Comparison of Tool Invocation Intensity. We Repeat the Experiment 4 Times and Compute the Mean and Std. Error bars Indicate Standard Deviation across 4 Times of Experiment.}
    \label{fig:tool_calls}
\end{figure}

\subsubsection{Analysis of Tool Call Number of Student Models}
To explore the use of tools, we count the exact number of queries searched and browsed per sample, as illustrated in Figure~\ref{fig:tool_calls}. 
We observe two key findings: (1) The original Qwen3 exhibits almost no multi-turn tool-calling capability, which explains its near-constant output of 0.5 for probability forecasting on ForecastBench and its lowest scores on L3--L4 of FutureX; (2) the \textit{Mix-SFT w/ harness} group does not consistently yield the highest number of queries, suggesting that the improved forecasting performance stems from the quality of query formulation and temporal reasoning rather than from test-time scaling via a larger number of tool calls.

\section{Discussion}

\subsection{Ablation Study}

\begin{table}[htbp]
\centering
\vspace{-10px}
\caption{Ablation Study on Components of Time-Truncation Harness.}
\label{tab:ablation}
\resizebox{0.9\textwidth}{!}{%
\begin{tabular}{lccccc}
\toprule
\textbf{Group} & 
\makecell[c]{\textbf{Time-Constrained}\\\textbf{Environment}} & 
\makecell[c]{\textbf{System Prompt}} & 
\makecell[c]{\textbf{Accuracy (Total)}} & 
\makecell[c]{\textbf{Average Tool Call}\\\textbf{Number (Total)}} & 
\makecell[c]{\textbf{Median Query}\\\textbf{Date-Span (Total)}} \\ \midrule
no harness    & $\times$                       & $\times$               & 83.61\%                   & 5.53                                      & 123                                     \\
prompt-only    & $\times$                       & $\checkmark$           & 84.61\%                   & 9.17                                      & 182                                     \\
env.-only  & $\checkmark$                       & $\times$               & 75.93\%                   & 8.09                                      & 365                                     \\
full harness   & $\checkmark$                   & $\checkmark$           & 74.27\%                   & 11.04                                     & 365                                     \\ \bottomrule
\end{tabular}
}
\end{table}

To evaluate the impact of environmental constraints and system prompts, we conduct an ablation study and the results are summarized in Table~\ref{tab:ablation}. The key findings are: (1) Introducing the time-constrained environment reduces accuracy from 84.61\% to 74.27\% (with prompt) and from 83.61\% to 75.93\% (without prompt), while expanding the median Query Date-Span from 182 to 365 days and from 123 to 365 days, respectively. This confirms that \textbf{the time-constrained environment is the primary mechanism}, imposing a hard boundary on the publication dates of retrievable documents. (2) \textbf{The system prompt plays a supporting role}: in the presence of the environment, it slightly increases the tool-call count, aiding temporal search and reasoning; without the environment, it raises both the tool-call count and the Query Date-Span (from 123 to 182 days), mildly reinforcing search behaviour, but cannot fundamentally resolve leakage, since the environment still permits access to post-cut-off documents.

\subsection{Sensitivity Analysis}
\begin{figure}
    \centering
    \vspace{-16px}
    \includegraphics[width=0.8\linewidth]
    {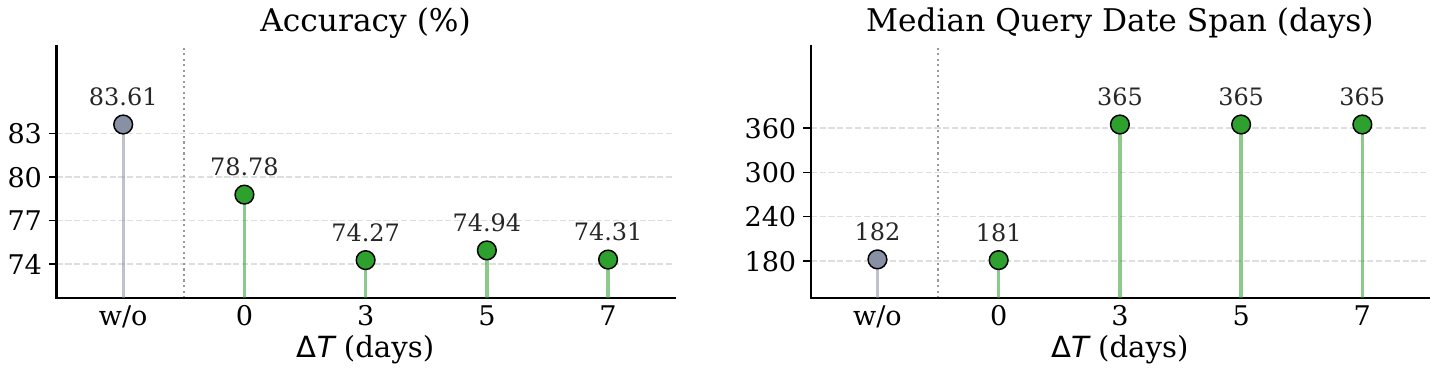}
    \caption{Sensitivity Analysis over $\Delta_T \in \{0,3,5,7\}$ Days.}
    \label{fig:sensitivity}
\end{figure}

To assess the influence of $\Delta_T$, we conduct a sensitivity analysis. As shown in Figure~\ref{fig:sensitivity}, when $\Delta_T = 0$, performance falls between that of the \textit{w/o-harness} group and that of the $\Delta_T \in \{3,5,7\}$ settings, indicating that the leakage filter is noticeably less reliable when the cut-off lies too close to the settlement date; in contrast, $\Delta_T = 3, 5, 7$ produce nearly identical accuracy and Query Date-Span, indicating that the temporal isolation becomes stable once a sufficient margin is imposed. We therefore select $\Delta_T = 3$ as the representative setting.

\subsection{Limitations and Future Works}

\paragraph{Limitations}
This work bears certain limitations: (1) The synthetic data quality remains fundamentally bounded by the inherent capabilities of the teacher model. (2) Our methodology focuses on analyzing data features and capability internalization of student models, rather than comparing with other baseline methods (which are hard to reproduce) or chasing SOTA performance (which still rely on agent framework). (3) While our harness improves data quality on average, it mitigates rather than fully eliminates temporal leakage, and lightweight rubric-based filtering could further reduce this fraction.

\paragraph{Future Works}
Future works include two primary directions: (1) We aim to scale the student model parameters to verify the observed improvement in forecasting capabilities persists. (2) We plan to repurpose the harness as a RL environment. Within this setup, the low-leakage feedback provided by the harness will be used to support cleaner rollouts and more reliable reward computation, enabling further exploration of optimized policies for long-trajectory temporal search.

\section{Conclusion}
\label{sec:conclusion}
In this paper, we present a time-truncation harness that synthesizes low-leakage forecasting trajectories from historical queries, substantially raising sampling efficiency and the proportion of high-quality temporal search and reasoning trajectories. Distillation experiments confirm that student models trained on these trajectories achieve better performance. Overall, this validates a paradigm for internalizing the external temporal search and reasoning behavior induced by our harness into the parametric capabilities of smaller models.

\newpage

\section*{Broader Impact}
\label{sec:broaderimpact}
The datasets used in this study (historical data from Manifold) are gathered entirely from publicly available platforms and open APIs, ensuring full compliance with data usage terms. To prevent the model from learning harmful content, we filtered out any queries or contexts containing toxic, violent, or explicitly biased information. All the data are solely for academic research purposes. Additionally, we emphasize that the future events predicted by LLM do not represent guaranteed representations of actual future occurrences. Misusing AI to place event prediction market bets may lead to financial losses, and blindly trusting AI predictions without human verification may cause unintended broader societal impacts.

\bibliographystyle{apalike}
\bibliography{references}

@article{miroflow,
  title={Miroflow: Towards high-performance and robust open-source agent framework for general deep research tasks},
  author={Shiqian Su and Sen Xing and Xuan Dong and Muyan Zhong and Bin Wang and Xizhou Zhu and Yuntao Chen and Wenhai Wang and Yue Deng and Pengxiang Zhu and Ziyuan Liu and Tiantong Li and Jiaheng Yu and Zhe Chen and Lidong Bing and Jifeng Dai},
  journal={arXiv preprint},
  volume={arXiv:2602.22808},
  year={2026}
}

@article{milkyway,
  title={Harnessing Pre-Resolution Signals for Future Prediction Agents},
  author={Chuyang Wei and Maohang Gao and Zhixin Han and Kefei Chen and Yu Zhuang and Haoxiang Guan and Yanzhi Zhang and Yilin Cheng and Xiren Zhou and Huanhuan Chen and Jian Li and Jiyan He and Yu Shi and Yitong Duan and Shuxin Zheng},
  journal={arXiv preprint},
  volume={arXiv:2604.15719},
  year={2026}
}

@article{flashsearcher,
  title={Flash-Searcher: Fast and Effective Web Agents via DAG-Based Parallel Execution},
  author={Tianrui Qin and Qianben Chen and Sinuo Wang and He Xing and King Zhu and He Zhu and Dingfeng Shi and Xinxin Liu and Ge Zhang and Jiaheng Liu and Yuchen Eleanor Jiang and Xitong Gao and Wangchunshu Zhou},
  journal={arXiv preprint},
  volume={arXiv:2509.25301},
  year={2025}
}

@misc{openclaw,
  author = {OpenClaw},
  title  = {OpenClaw},
  year   = {2026},
  howpublished = {\url{https://github.com/openclaw/openclaw}}
}

@misc{claudecode,
  author = {Anthropic},
  title  = {Claude Code | Anthropic's agentic coding system},
  year   = {2025},
  howpublished  = {\url{https://www.anthropic.com/product/claude-code}}
}

@misc{codex,
  author = {OpenAI},
  title  = {Introducing Codex},
  year   = {2025},
  howpublished    = {\url{https://openai.com/index/introducing-codex}}
}

@misc{openaideepresearch,
  author       = {OpenAI},
  title        = {OpenAI Deep Research: Advanced iterative search and agentic synthesis},
  year         = {2025},
  howpublished = {\url{https://openai.com/index/introducing-deep-research}}
}

@misc{geminideepresearch,
  author       = {{Google DeepMind}},
  title        = {Build with Gemini Deep Research},
  year         = {2025},
  howpublished = {\url{https://blog.google/innovation-and-ai/technology/developers-tools/deep-research-agent-gemini-api}}
}

@misc{h2o,
  author       = {h2o.ai},
  title        = {What Makes H2O AI Super Agent™ Super?},
  year         = {2026},
  howpublished = {\url{https://h2o.ai/blog/2026/what-makes-h2oai-super-agent-super}}
}

@article{glm5,
      title={GLM-5: from Vibe Coding to Agentic Engineering}, 
      author={GLM-Team and others},
      year={2026},
      journal={arXiv preprint},
      volume={arXiv:2602.15763}, 
}

@article{kimik2.5,
  title={Kimi K2.5: Visual Agentic Intelligence},
  author={Kimi-Team and others},
  journal={arXiv preprint},
  volume={arXiv:2602.02276},
  year={2026}
}

@article{longcatflashthinking2601,
  title={LongCat-Flash-Thinking-2601 Technical Report},
  author={LongCat-Team and others},
  journal={arXiv preprint},
  volume = {arXiv:2601.16725},
  year={2026}
}

@article{react,
  author    = {Yao, Shunyu and Zhao, Jeffrey and Yu, Dian and Du, Nan and Shafran, Izhak and Narasimhan, Karthik and Cao, Yuan},
  title     = {ReAct: Synergizing Reasoning and Acting in Language Models},
  journal   = {arXiv preprint},
  year      = {2022},
  volume    = {arXiv:2210.03629}
}

@article{searchr1,
  author    = {Jin, Bowen and Zeng, Hansi and Yue, Zhenrui and Yoon, Jinsung and Arik, Sercan and Wang, Dong and Zamani, Hamed and Han, Jiawei},
  title     = {Search-R1: Training LLMs to Reason and Leverage Search Engines with Reinforcement Learning},
  journal   = {arXiv preprint},
  year      = {2025},
  volume    = {arXiv:2503.09516}
}

@article{deepdive,
  author    = {Lu, Rui and Hou, Zhenyu and Wang, Zihan and Zhang, Hanchen and Liu, Xiao and Li, Yujiang and Feng, Shi and Tang, Jie and Dong, Yuxiao},
  title     = {DeepDive: Advancing Deep Search Agents with Knowledge Graphs and Multi-Turn RL},
  journal   = {arXiv preprint},
  year      = {2025},
  volume    = {arXiv:2509.10446}
}

@article{mirothinkerv1.7h1,
  title={MiroThinker-1.7 \& H1: Towards Heavy-Duty Research Agents via Verification}, 
      author={MiroMind-Team and S. Bai and L. Bing and L. Lei and R. Li and X. Li and X. Lin and E. Min and L. Su and B. Wang and L. Wang and L. Wang and S. Wang and X. Wang and Y. Zhang and Z. Zhang and G. Chen and L. Chen and Z. Cheng and Y. Deng and Z. Huang and D. Ng and J. Ni and Q. Ren and X. Tang and B. L. Wang and H. Wang and N. Wang and C. Wei and Q. Wu and J. Xia and Y. Xiao and H. Xu and X. Xu and C. Xue and Z. Yang and Z. Yang and F. Ye and H. Ye and J. Yu and C. Zhang and W. Zhang and H. Zhao and P. Zhu},
  journal   = {arXiv preprint},
  year      = {2026},
  volume    = {arXiv:2603.15726}
}

@misc{echo,
  title   = {Echo: Towards General AI Prediction},
  author  = {UniPat-AI},
  year    = {2026},
  howpublished     = {\url{https://unipat.ai/blog/Echo}}
}

@article{prophet,
  author    = {Tao, Zhengwei and Jin, Zhi and Li, Bincheng and Bai, Xiaoying and Zhao, Haiyan and Dou, Chengfeng and Chen, Xiancai and Li, Jia and Li, Linyu and Tao, Chongyang},
  title     = {PROPHET: An Inferable Future Forecasting Benchmark with Causal Intervened Likelihood Estimation},
  journal   = {arXiv preprint},
  year      = {2025},
  volume    = {arXiv:2504.01509}
}

@article{llmasprophet,
  title={LLM-as-a-Prophet: Understanding Predictive Intelligence with Prophet Arena},
  author={Yang, Qingchuan and Mahns, Simon and Li, Sida and Gu, Anri and Wu, Jibang and Xu, Haifeng},
  journal={arXiv preprint},
  volume={arXiv:2510.17638},
  year={2025}
}

@article{forecastbench,
  author    = {Karger, Ezra and Bastani, Houtan and Chen, Yueh-Han and Jacobs, Zachary and Halawi, Danny and Zhang, Fred and Tetlock, Philip E.},
  title     = {ForecastBench: A Dynamic Benchmark of AI Forecasting Capabilities},
  journal   = {arXiv preprint},
  year      = {2025},
  volume    = {arXiv:2409.19839}
}

@article{futurex,
  author    = {Zeng, Zhiyuan and Liu, Jiashuo and Chen, Siyuan and He, Tianci and Liao, Yali and Tian, Yixiao and Wang, Jinpeng and Wang, Zaiyuan and Yang, Yang and Yin, Lingyue and Yin, Mingren and Zhu, Zhenwei and Cai, Tianle and Chen, Zehui and Chen, Jiecao and Du, Yantao and Gao, Xiang and Guo, Jiacheng and Hu, Liang and Jiao, Jianpeng and Li, Xiangsheng and Liu, Jingkai and Ni, Shuang and Wen, Zhoufutu and Zhang, Ge and Zhang, Kaiyuan and Zhou, Xin and Blanchet, Jose and Qiu, Xipeng and Wang, Mengdi and Huang, Wenhao},
  title     = {FutureX: An Advanced Live Benchmark for LLM Agents in Future Prediction},
  journal   = {arXiv preprint},
  year      = {2025},
  volume    = {arXiv:2508.11987}
}

@article{futurexpro,
  author    = {Liu, Jiashuo and Chen, Siyuan and Wang, Zaiyuan and Zeng, Zhiyuan and Guo, Jiacheng and Hu, Liang and Yin, Lingyue and Huang, Suozhi and Hao, Wenxin and Yang, Yang and Cheng, Zerui and Yao, Zixin and Yin, Lingyue and Liu, Haoxin and Cheng, Jiayi and Li, Yuzhen and Ma, Zezhong and Wang, Bingjie and Qiu, Bingsen and Liu, Xiao and Zhang, Zeyang and Liu, Zijian and Wang, Jinpeng and Yin, Mingren and He, Tianci and Liao, Yali and Tian, Yixiao and Zhu, Zhenwei and Dai, Anqi and Zhang, Ge and Liu, Jingkai and Zhang, Kaiyuan and Wu, Wenlong and Gao, Xiang and Chen, Xinjie and Yao, Zhixin and Wen, Zhoufutu and Prakash, B. Aditya and Blanchet, Jose and Wang, Mengdi and Si, Nian and Huang, Wenhao},
  title     = {FutureX-Pro: Extending Future Prediction to High-Value Vertical Domains},
  journal   = {arXiv preprint},
  year      = {2026},
  volume    = {arXiv:2601.12259}
}

@article{metaculus,
  author    = {{Metaculus}},
  title     = {AI Forecasting Benchmark Series Q2 (2025)},
  journal   = {Metaculus},
  year      = {2025}
}

@article{forecastqa,
  author    = {Jin, Woojeong and Khanna, Rahul and Kim, Suji and Lee, Dong-Ho and Morstatter, Fred and Galstyan, Aram and Ren, Xiang},
  title     = {ForecastQA: A Question Answering Challenge for Event Forecasting with Temporal Text Data},
  journal   = {arXiv preprint},
  year      = {2021},
  volume    = {arXiv:2005.00792}
}

@article{autocast,
  author    = {Zou, Andy and Xiao, Tristan and Jia, Ryan and Kwon, Joe and Mazeika, Mantas and Li, Richard and Song, Dawn and Steinhardt, Jacob and Evans, Owain and Hendrycks, Dan},
  title     = {Forecasting Future World Events with Neural Networks},
  journal   = {arXiv preprint},
  year      = {2022},
  volume    = {arXiv:2206.15474}
}

@article{futureworld,
  author    = {Han, Zhixin and Zhang, Yanzhi and Wei, Chuyang and Gao, Maohang and Yue, Xiawei and Chen, Kefei and Zhuang, Yu and Guan, Haoxiang and He, Jiyan and Li, Jian and Duan, Yitong and Shi, Yu and Hu, Mengting and Zheng, Shuxin},
  title     = {FutureWorld: A Live Reinforcement Learning Environment for Predictive Agents with Real-World Outcome Rewards},
  journal   = {arXiv preprint},
  year      = {2026},
  volume    = {arXiv:2604.26733}
}

@article{finance,
  title={A comprehensive look at the empirical performance of equity premium prediction},
  author={Welch, Ivo and Goyal, Amit},
  journal={The Review of Financial Studies},
  volume={21},
  number={4},
  pages={1455--1508},
  year={2008},
  publisher={Society for Financial Studies}
}

@article{publichealth,
  title={Evaluation of individual and ensemble probabilistic forecasts of COVID-19 mortality in the United States},
  author={Cramer, Estee Y and Ray, Evan L and Lopez, Velma K and Bracher, Johannes and Brennen, Andrea and Castro Rivadeneira, Alvaro J and Gerding, Aaron and Gneiting, Tilmann and House, Katie H and Huang, Yuxin and others},
  journal={Proceedings of the National Academy of Sciences},
  volume={119},
  number={15},
  pages={e2113561119},
  year={2022},
  publisher={National Academy of Sciences}
}

@article{politic,
  title={Psychological strategies for winning a geopolitical forecasting tournament},
  author={Mellers, Barbara and Ungar, Lyle and Baron, Jonathan and Ramos, Jaime and Gurcay, Burcu and Fincher, Katrina and Scott, Sydney E and Moore, Don and Atanasov, Pavel and Swift, Samuel A and others},
  journal={Psychological science},
  volume={25},
  number={5},
  pages={1106--1115},
  year={2014},
  publisher={Sage Publications Sage CA: Los Angeles, CA}
}

@article{approaching,
  title={Approaching human-level forecasting with language models},
  author={Halawi, Danny and Zhang, Fred and Yueh-Han, Chen and Steinhardt, Jacob},
  journal={Advances in Neural Information Processing Systems},
  volume={37},
  pages={50426--50468},
  year={2024}
}

@article{outcome,
  title={Outcome-based Reinforcement Learning to Predict the Future},
  author={Turtel, Benjamin and Franklin, Danny and Skotheim, Kris and Hewitt, Luke and Schoenegger, Philipp},
  journal   = {arXiv preprint},
  volume    = {arXiv:2505.17989},
  year={2025}
}

@article{futureaslabel,
  title={Future-as-Label: Scalable Supervision from Real-World Outcomes},
  author={Turtel, Benjamin and Wilczewski, Paul and Franklin, Danny and Skothiem, Kris},
  journal={arXiv preprint},
  volume = {arXiv:2601.06336},
  year={2026}
}

@misc{thinkingmatchines,
  author = {Scott, Jeen and Matthew, Aitchison and Mantic},
  title = {Training LLMs to Predict World Events},
  journal = {Thinking Machines Lab: News},
  year = {2026},
  howpublished = {\url{https://thinkingmachines.ai/news/training-llms-to-predict-world-events}}
}

@article{browsecomp,
  title={Browsecomp: A simple yet challenging benchmark for browsing agents},
  author={Wei, Jason and Sun, Zhiqing and Papay, Spencer and McKinney, Scott and Han, Jeffrey and Fulford, Isa and Chung, Hyung Won and Passos, Alex Tachard and Fedus, William and Glaese, Amelia},
  journal={arXiv preprint},
  volume = {arXiv:2504.12516},
  year={2025}
}

@article{browsecompzh,
  title={Browsecomp-zh: Benchmarking web browsing ability of large language models in chinese},
  author={Zhou, Peilin and Leon, Bruce and Ying, Xiang and Zhang, Can and Shao, Yifan and Ye, Qichen and Chong, Dading and Jin, Zhiling and Xie, Chenxuan and Cao, Meng and others},
  journal={arXiv preprint},
  volume={arXiv:2504.19314},
  year={2025}
}

@misc{qwen3,
      title={Qwen3 Technical Report}, 
      author={Qwen-Team and others},
      year={2025},
      volume={arxiv:2505.09388},
      journal={arXiv}
}

@inproceedings{vllm,
  title={Efficient Memory Management for Large Language Model Serving with PagedAttention},
  author={Woosuk Kwon and Zhuohan Li and Siyuan Zhuang and Ying Sheng and Lianmin Zheng and Cody Hao Yu and Joseph E. Gonzalez and Hao Zhang and Ion Stoica},
  booktitle={Proceedings of the ACM SIGOPS 29th Symposium on Operating Systems Principles},
  year={2023}
}

@article{tir1,
  title={Teaching language models to reason with tools},
  author={Li, Chengpeng and Tang, Zhengyang and Li, Ziniu and Xue, Mingfeng and Bao, Keqin and Ding, Tian and Sun, Ruoyu and Wang, Benyou and Wang, Xiang and Lin, Junyang and others},
  journal={Advances in Neural Information Processing Systems},
  volume={38},
  pages={66843--66892},
  year={2026}
}

@inproceedings{tir2,
  title={TInR: Exploring Tool-Internalized Reasoning in Large Language Models},
  author={Xu, Qiancheng and Li, Yongqi and Liu, Fan and Wang, Hongru and Yang, Min and Li, Wenjie},
  booktitle={Proceedings of the 64th Annual Meeting of the Association for Computational Linguistics (Volume 1: Long Papers)},
  pages={44851--44865},
  year={2026}
}

@misc{claudeopus4.8,
  author = {Anthropic},
  title  = {Introducing Claude-opus-4.8},
  year   = {2026},
  howpublished = {\url{https://www.anthropic.com/news/claude-opus-4-8}}
}

\newpage
\appendix
\section*{Appendix}

\begingroup 
\color{black}
\hypersetup{linkcolor=black}
\startcontents[appendices]
\printcontents[appendices]{l}{1}{\setcounter{tocdepth}{2}}
\endgroup

\section{More Related Works}
\label{appendix:benchmarks}

\paragraph{Benchmarks for Future Prediction}
Evaluating future prediction agents requires benchmarks that genuinely assess predictive capacity rather than recall. Early benchmarks such as ForecastQA \citep{forecastqa}, Autocast \citep{autocast} and static Metaculus collections \citep{metaculus} are built from historical queries with known outcomes; since these questions date back to around 2022, they are susceptible to memorization and primarily test retrieval rather than generalization to new events. PROPHET \citep{prophet} and LLM-as-a-prophet \citep{llmasprophet} provide each query with a fixed set of pre-collected evidence, which supports controlled evaluation of reasoning over given documents but does not accommodate agent-driven search or direct comparison across varying retrieval trajectories. More recent efforts, including FutureX \citep{futurex} and ForecastBench \citep{forecastbench}, adopt dynamically updated question sets with explicit resolution dates and ground-truth answers, enabling evaluation on current, unseen events while allowing agents to perform their own evidence gathering. FutureX-Pro \citep{futurexpro} also follows a dynamic updated schedule and targets verifiable domain-specific topics, allowing more reliable evaluation of both prediction processes and results.

\section{Methodology Details}

\subsection{Tool Schema of Search Agent}
\label{appendix:tools}

\begin{guiguardprompt}{Tool Schema}
[
  {
    "type": "function",
    "function": {
      "name": "search",
      "description": "General-purpose web search. Return top results with relevant snippets.",
      "parameters": {
        "type": "object",
        "required": ["query"],
        "properties": {
          "query": {
            "type": "array",
            "description": "Array of search queries (max 5). Executed in parallel within one step.",
            "items": { "type": "string" }
          }
        }
      }
    }
  },
  {
    "type": "function",
    "function": {
      "name": "browse_with_goal",
      "description": "Opens specific URL and extracts content relevant to a given goal. Instead of returning the full page, only goal-related content is returned, which saves context and improves accuracy.",
      "parameters": {
        "type": "object",
        "required": ["url", "goal"],
        "properties": {
          "url": {
            "type": "array",
            "description": "URL to fetch.",
            "items": { "type": "string" }
          },
          "goal": {
            "type": "string",
            "description": "The specific information you want to extract from the page. e.g. \"Find the release date of product X\" or \"Extract the financial data for Q3 2024\"."
          }
        }
      }
    }
  }
]
\end{guiguardprompt}

\subsection{System Prompts for Time-Truncation Harness}
\label{appendix:system prompt}

\begin{guiguardprompt}{System Prompt $S_0$}
Today is T_now. Your responsibility is to use all available tools to collect, summarize, and organize as much relevant information as possible to answer the user's question.
Notice:
- This event or fact has already occurred or is currently unfolding relative to T_now. You need to search for the event and verify any details."
- Do NOT rely on your prior training knowledge to draw conclusions. Treat your internal knowledge as potentially outdated, incomplete, or biased. All reasoning must be strictly grounded in information retrieved through your search and browsing tools.
- Do NOT ask the user any questions under any circumstances. You must independently resolve ambiguities and drive the research process. You may briefly reflect before taking action.
\end{guiguardprompt}

\begin{guiguardprompt}{Modified System Prompt $\tilde{S_0}$}
Today is T_cut. Your task is to predict the outcome of future events.
Critical Premises
- This event has not yet occurred. There is no correct answer to be found online.
- Do NOT rely on your prior training knowledge to draw conclusions. Treat your internal knowledge as potentially outdated or biased. All reasoning must be grounded in information retrieved through your search and browsing tools.
Research Process
Make Plan
- You **MUST** first generate a query plan, which includes all the information you need to know in order to predict this event. This **MUST** include:
  - background, relevant data and events.
  - positive/negative/multi-aspect factors that have had a significant impact on the target event in the recent past.
  - views from the markets and human experts.
Information Gathering
- Use all the tools to complete your Plan strictly.
- Your search queries should be diverse and well-targeted.
Reasoning After Search
- After each round of tool calls, reason strictly based on information you have collected.
- For each piece of information retrieved, explicitly analyze:
  - **Temporal proximity**: How close in time is this information to the predicted event?
  - **Source credibility**: How reliable and authoritative is the source?
  - **Directional impact**: How do these factors influence the target event that you are trying to predict?
Iteration & Stopping Condition
- If your confidence is insufficient to make a well-supported prediction, identify what additional information you still need and continue searching.
- Only stop and deliver your final prediction when you are confident that you have gathered enough evidence to support a reasoned conclusion.
\end{guiguardprompt}

\section{Experimental Details}

\subsection{Manifold Query Collection}
\label{appendix:data}
\paragraph{Fields and Types} We enumerate 30 topic slugs by \texttt{GET /v0/groups}, spanning 
\texttt{technology-default}, \texttt{science-default}, \texttt{economics-default},
\texttt{ai}, \texttt{politics-default}, \texttt{us-politics}, \texttt{culture-default},
\texttt{world-default}, \texttt{space}, \texttt{sports-default}, \texttt{gaming},
\texttt{finance}, \texttt{programming}, \texttt{climate}, \texttt{movies},
\texttt{crypto-speculation}, \texttt{health}, \texttt{stocks}, \texttt{mathematics},
\texttt{europe}, \texttt{china}, \texttt{geopolitics}, \texttt{middle-east},
\texttt{india}, \texttt{internet}, \texttt{russia}, \texttt{celebrities},
\texttt{business}, \texttt{biotech}, and \texttt{energy}. For each slug, we issue paginated requests to the \texttt{GET /search-markets} endpoint with
\texttt{filter=resolved}, \texttt{contractType=ALL}, \texttt{sort=newest}, and a
batch size of 1{,}000. Pagination walks backward in time: after each batch, the
\texttt{createdTime} of the last returned market becomes the \texttt{beforeTime}
anchor of the subsequent request. Collection for a slug terminates once 2{,}000
qualified markets have been accumulated or after three consecutive empty
batches.

\paragraph{Verifiability and Objectivity} Each returned market is retained only if it passes a sequence of filters
enforcing verifiability and objectivity. We first require that the market be
already resolved (\texttt{isResolved=True}) and that its \texttt{createdTime}
and \texttt{resolutionTime} parse to valid dates. Markets resolved to the market price (\texttt{MKT}) or cancelled
(\texttt{CANCEL}) are dropped since these carry no objective ground truth.
We then
discard personal or subjective questions whose text contains any
first-person marker (\texttt{will i}, \texttt{i will}, \texttt{should i},
\texttt{am i}, \texttt{i am}, \texttt{i'm}, \texttt{my}, or \texttt{me}). Finally, we keep
only markets whose \texttt{outcomeType} is \texttt{Binary}, \texttt{Multiple-choice}, \texttt{Date}, or
\texttt{Numeric}, dropping \texttt{Opinions} and \texttt{Others}.

\paragraph{Answer Extraction} Answer extraction depends on the market type. For binary markets, the choice set
is fixed to \{Yes, No\} and each question has an answer in \texttt{Resolution} field and a probability distribution $[p,1-p]$ for Yes and No. For
multiple-choice, date, and numeric markets, we issue a second request to
\texttt{GET /market/\{id\}} to obtain the full \texttt{answers} list, from which
we take each answer's text as a choice and its probability distribution as the corresponding
choice probability; the ground-truth label is the answer whose identifier matches
the market's \texttt{Resolution} field, together with its probability. Every
retained market is serialized as a single JSONL record containing its \texttt{Identifier},
\texttt{URL}, \texttt{Question}, \texttt{Type}, \texttt{Creation Time} and \texttt{Resolution Time} (formatted as
\texttt{YYYY-MM-DD}), the list of \texttt{Choices} and the \texttt{Resolution} with its \texttt{Probability}.

\paragraph{Temporal Filter} The raw crawl is then refined by an offline post-processing steps. We apply a temporal filter that retains only markets resolving between June 2025 and March 2026, thereby eliminating the risk of pre-training contamination in the student models and of overlap with the benchmark evaluation window.

\subsection{Trajectory Rollout Settings}
\label{appendix:teacher_config}

\begin{table}[h]
\centering
\caption{Rollout Parameter Settings for Teacher Model}
\label{tab:sampling_params}
\small
\begin{tabular}{lc}
\hline
\textbf{Parameter} & \textbf{Value} \\ \hline
TOP\_P              & 1.0            \\
TEMP (Temperature)  & 1.0            \\
FREQUENCY\_PENALTY  & 0              \\
TOP\_K              & 20             \\
PRESENCE\_PENALTY   & 0              \\
REPETITION\_PENALTY & 1.0            \\ \hline
\end{tabular}
\end{table}

\paragraph{Configuration} We use the official Kimi-K2.5 API provided by Moonshot AI to synthesize trajectories. The configuration settings are summarized in Table \ref{tab:sampling_params}. The aim is to introduce a moderate degree of randomness while filtering out highly improbable tokens, which prevents the generated trajectories from becoming overly repetitive or rigid. By maintaining high exploration thresholds and bounding the token pool, it encourages diverse reasoning paths to fully leverage the teacher model's intrinsic capabilities and behavioral patterns.

\paragraph{Context Management} We adhere to Kimi-K2.5's default context limit of 128K tokens without employing any context management techniques (such as summary, discard all, or other context compression method), thereby retaining the raw, uncompressed trajectories and preserving their behavioural features.

\subsection{Content Leakage Verification Prompt}
\label{appendix:verify prompt}

\begin{guiguardprompt}{Verification Prompt}
You are a professional temporal leakage verifier. Given a query, a cutoff time T_cut, and all the content returned by a search engine, for each entry with datePublished < T_cut, determine whether the returned content contains a direct description of events occurring at t >= T_cut.
You must only return: [YES] or [NO]
\end{guiguardprompt}

\subsection{Query Date-Span Computation}
\label{appendix:querydatespan}

\paragraph{Date Parser}
We extract and parse time-related text from each query written by the agent in each tool calls (\texttt{search, browse with goal}), according to the following priority:
\begin{itemize}
  \item ISO / numeric dates: \texttt{YYYY-MM-DD} and \texttt{YYYY/MM/DD};
  \item day--month--year: \texttt{17 May 2026}, \texttt{17th of May 2026};
  \item month--day--year: \texttt{May 17 2026}, \texttt{May 17, 2026};
  \item month--year: \texttt{May 2026};
  \item year--month: \texttt{YYYY-MM};
  \item fiscal periods: \texttt{Q1 2026}, \texttt{H2 2025};
  \item bare four-digit years: \texttt{2024}, \texttt{2025}, \texttt{2026}.
\end{itemize}

\paragraph{Overlap Resolution and Priority}
Because several patterns can match overlapping substrings (e.g., the year inside a
full ISO date), matches are accepted greedily in the priority order above and
deduplicated by character span: once a span is consumed, any later match that
overlaps it is discarded. The bare-year pattern is applied last so that a full date
is never reduced to just its year.

\paragraph{Completion of Under-specified Dates}
Partial dates are completed to a fixed representative day so that they can enter the
min/max computation as ordinary calendar dates: a month--year maps to the first of
that month; a bare year maps to July~1; a fiscal quarter \texttt{Q1--Q4} maps to the
$15$th of a representative month (February, May, August, November); and a half-year
\texttt{H1}/\texttt{H2} maps to the $15$th of March / September. These choices only
shift a date within its own period and therefore have negligible effect on
day-scale spans.

\paragraph{Relative Expressions}
Purely relative time expressions (include \texttt{"today", "tonight", "yesterday",
"past~/~last~/~next $N$ days\,/\,weeks\,/\,months\,/\,years", "year-to-date",
"as of \dots"}) are detected separately and are \textbf{NOT} converted to concrete
dates. They do not contribute to the Query Date-Span, since such expressions do not resolve to concrete dates in search engines.

\subsection{SFT Settings}
\label{appendix:sft}
\paragraph{Trajectory Format Processing}
We enforce a strict alignment protocol across all SFT data trajectories. The formatting specifications are structured as follows:

\begin{itemize}
    \item \textbf{Unified System Prompt:} The system prompt is entirely standardized across all instances, serving three critical objectives: 
    (i) to seamlessly adapt to the newly introduced \texttt{<final\_summary>} formatting constraint; 
    (ii) to significantly economize token consumption given the strict 32K token context limit of the \texttt{Qwen3} architecture; and 
    (iii) to prevent substantial SFT performance deviations induced by excessive variances in system prompts, thereby ensuring a rigorous and fair comparison during the TIR processes.
    \begin{guiguardprompt}{Unified System Prompt for SFT and Inference}
Today is T_cut (or T_now). Your responsibility is to use all available tools to collect, summarize, and organize as much relevant information as possible to answer the user's question. Before providing an answer, carefully verify details from multiple perspectives to ensure your response is accurate and comprehensive. Do not ask the user any questions. You may briefly reflect before taking action. Enclose your final response within the <final_summary> and </final_summary> tags.
\end{guiguardprompt}
    \item \textbf{Token Enclosure Protocol:} Following the standard Qwen3 SFT configuration, structural tags are utilized to segment conversational turns and operational paradigms. Specifically, human inputs are encapsulated via \texttt{<|im\_start|>user\dots<|im\_end|>}, while agent responses are bounded by \texttt{<|im\_start|>assistant\dots<|im\_end|>}.
    The internal reasoning process of the model is enclosed within explicit \texttt{<think>\dots</think>} tags. External function execution invocations and their corresponding execution feedbacks are dynamically wrapped using \texttt{<tool\_call>\dots</tool\_call>} and \texttt{<tool\_result>\dots</tool\_result>} tags, respectively.
    \item \textbf{Final Answer Standardization:} To ensure complete alignment across diverse structural corpora, the conversational segment where the agent formulates its final summary and ultimate response to the user is tightly wrapped within \texttt{<final\_summary>\dots</final\_summary>} tags.
\end{itemize}

\paragraph{Hyperparameters and Hardwares}
We perform full-parameter fine-tuning on a cluster comprising $512 \times \text{Huawei Ascend 910B (64GB)}$ NPUs. The hyperparameters and parallel strategies for both the 8B and 32B model scales are outlined in Table~\ref{tab:sft_configs}.

\begin{table}[htbp]
\centering
\caption{Hyperparameters and Parallelization Configurations for Qwen3-8B and Qwen3-32B Full-Parameter SFT.}
\label{tab:sft_configs}
\small
\vspace{6px}
\begin{tabular}{llcc}
\hline
\textbf{Category} & \textbf{Hyperparameter} & \textbf{Qwen3-8B} & \textbf{Qwen3-32B} \\ \hline
\multirow{3}{*}{Parallelism Strategies} & Tensor Parallel Size (TP) & 8 & 8 \\
 & Pipeline Parallel Size (PP) & 6 & 8 \\
 & Context Parallel Size (CP) & 4 & 8 \\ \hline
\multirow{3}{*}{Batch Settings} & Global Batch Size & 64 & 128 \\
 & Micro Batch Size & 1 & 1 \\
 & Sequence Length & 65,536 & 65,536 \\ \hline
Training Schedule & Train Epochs & 4 & 3 \\ \hline
\multirow{4}{*}{Learning Rate Schedule} & Learning Rate (LR) & $4\times 10^{-5}$ & $2\times 10^{-5}$ \\
 & Min Learning Rate & $1\times 10^{-5}$ & $1\times 10^{-7}$ \\
 & LR Decay Style & Cosine & Cosine \\
 & LR Warmup Fraction & 0.1 & 0.1 \\ \hline
\multirow{3}{*}{Optimizer Settings} & Adam $\beta_1$ / $\beta_2$ & 0.9 / 0.95 & 0.9 / 0.95 \\
 & Weight Decay & 0 & $1\times 10^{-2}$ \\
 & Clip Gradient & 1.0 & 1.0 \\ \hline
Regularization & Attention Dropout & 0 & 0 \\ \hline
\multirow{2}{*}{Precision \& Memory} & Mixed Precision & BF16 (\checkmark) & BF16 (\checkmark) \\
 & Recompute Granularity & Selective & Selective \\ \hline
\end{tabular}
\end{table}

\paragraph{SFT Loss with Selective Tool Masking}
We use the standard token-masking strategy for TIR during full-parameter SFT. Given an autoregressive sequence of tokens $X = \{x_1, x_2, \dots, x_N\}$, the localized optimization object focuses strictly on the generative reasoning capabilities of the agent.

Let $M(i) \in \{0, 1\}$ denote a binary supervisor mask for token $x_i$ at position $i$. The indicator function is defined explicitly based on the contextual blocks as follows:
\begin{equation}
M(i) = \begin{cases} 
1, & \text{if } x_i \in \mathcal{S}_{\text{think}} \cup \mathcal{S}_{\text{tool\_call}} \cup \mathcal{S}_{\text{final\_summary}} \\
0, & \text{if } x_i \in \mathcal{S}_{\text{tool\_result}}
\end{cases}
\end{equation}
where $\mathcal{S}_{\text{think}}$, $\mathcal{S}_{\text{tool\_call}}$, and $\mathcal{S}_{\text{final\_summary}}$ represent the sets of tokens enclosed within the \texttt{<think>}, \texttt{<tool\_call>}, and \texttt{<final\_summary>} tags respectively, whereas tokens originating from the external tool observation (\texttt{<tool\_result>}) are explicitly masked out.

The loss function is formalized via the cross-entropy loss over the masked sequence:
\begin{equation}
\mathcal{L}_{\text{SFT}}(\theta) = - \frac{1}{\sum_{i=1}^{N} M(i)} \sum_{i=1}^{N} M(i) \cdot \log P_\theta \left( x_i \mid x_{<i} \right)
\end{equation}
where $\theta$ represents the trainable parameters of the language model, and $P_\theta(x_i \mid x_{<i})$ is the predictive probability of token $x_i$ conditioned on the historical prefix context $x_{<i}$.

\begin{figure}
    \centering
    \includegraphics[width=\linewidth]{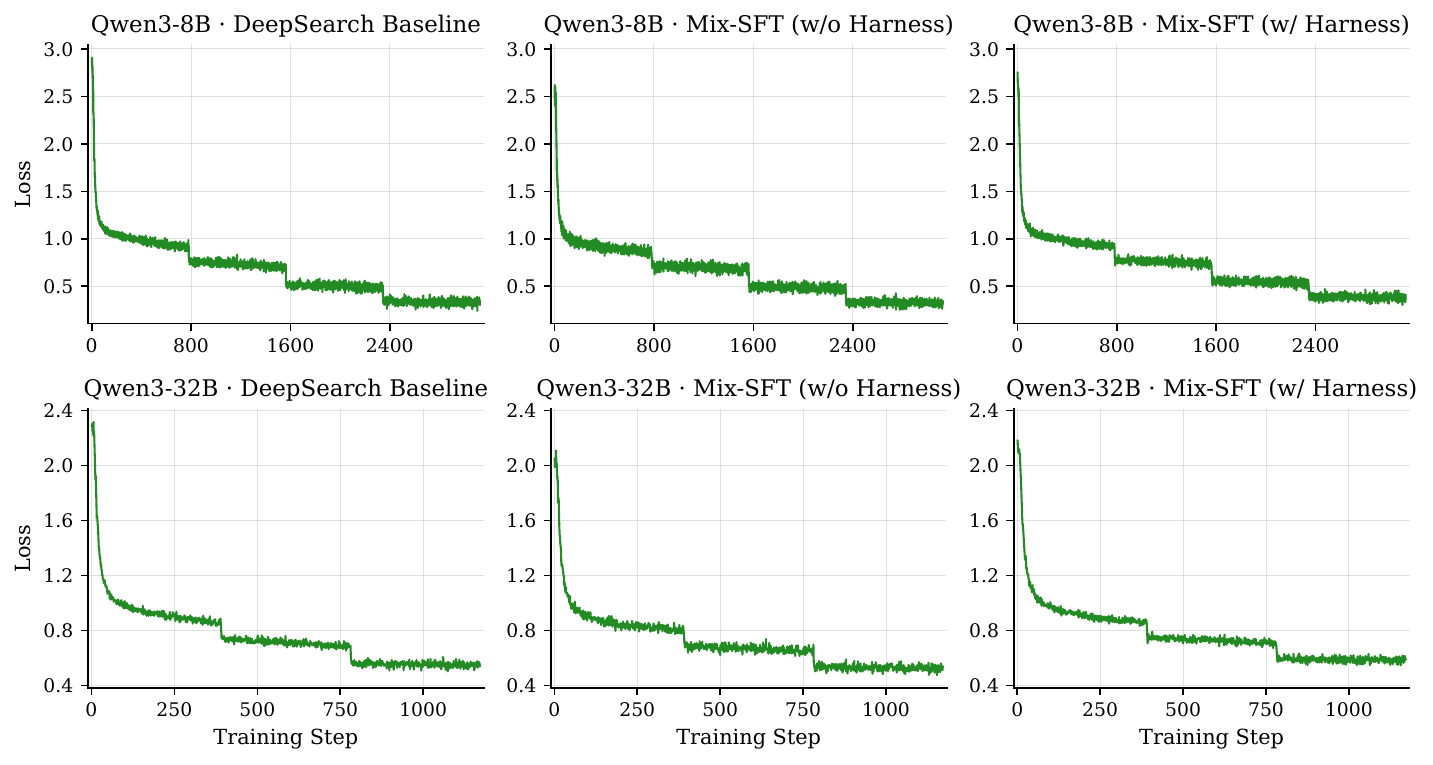}
    \caption{The SFT Loss Curve of Qwen3-8B and 32B}
    \label{fig:sft_loss_curve}
\end{figure}

\paragraph{SFT Loss Curve} The convergence trajectory of $\mathcal{L}_{\text{SFT}}$ is illustrated in Figure~\ref{fig:sft_loss_curve}. As empirically observed from the curve, the loss exhibits a distinct, ladder-like stepwise descent across subsequent iterations. In the initial iterations, the loss drops rapidly, demonstrating highly efficient knowledge acquisition. Throughout the final complete iteration, the token-level loss smoothly plateaus and stabilizes on a per-sample basis. This training dynamic suggests that the model assimilates the complex multi-turn execution trajectories while the risk of overfitting remains controlled.

\subsection{Benchmarks and Metrics}
\label{appendix:benchmarksettings}

\paragraph{ForecastBench}
ForecastBench spans diverse domains, aggregating data from 8 authoritative real-world sources:
\begin{itemize}
    \item \textbf{ACLED:} Tracking geopolitical conflicts, regional violence, and strategic instability events globally.
    \item \textbf{DBnomics:} Providing high-coverage global macroeconomic datasets.
    \item \textbf{FRED:} Utilizing Federal Reserve Economic Data to monitor macroeconomic indicators and financial trends.
    \item \textbf{Wikipedia:} Incorporating evolving encyclopedic articles and dynamic structured information.
    \item \textbf{Yahoo Finance:} Extracting real-time and historical financial market equity prices.
    \item \textbf{Manifold:} A user-generated prediction platform tracking public sentiment across various topics.
    \item \textbf{Metaculus:} A reputation-based forecasting platform focused on science, technology, and global trends.
    \item \textbf{Polymarket:} A decentralized prediction market capturing real-time probability data on global events.
    
\end{itemize}

\paragraph{Data Splitting Strategy} For ForecastBench, we construct the evaluation set using the \textit{May Slice}. The pre-training and SFT corpus is strictly frozen behind March 2026. While alternative prediction markets were initially considered, platforms such as Manifold (8 available queries), Polymarket (17 available queries), and Metaculus (2 available queries) were discarded due to severe sample sparsity and a high risk of being In-Distribution with respect to the SFT corpus. Conversely, the five selected data sources offer abundant evaluation instances and represent robust Out-of-Distribution (OOD) scenarios. 

\paragraph{Evaluation Settings} Following standard practices, the evaluation timeline is aligned with the original benchmark setup by setting the information cutoff time as $T_{\text{cut}} = \texttt{Freeze Time}$.

\paragraph{Metric} For binary forecasting questions, we report the \textbf{Brier Score} defined as:
\begin{equation}
\mathcal{B} = \frac{1}{N} \sum_{t=1}^{N} (f_t - o_t)^2
\end{equation}
where $N$ denotes the total number of evaluation instances, $f_t \in [0, 1]$ represents the predicted forecast probability assigned to the occurrence of the target event, and $o_t \in \{0, 1\}$ indicates the binary ground-truth outcome.

\paragraph{FutureX}
FutureX targets highly challenging, short-to-mid-term future event horizons. To mitigate temporal instability and ensure robust signal estimation, we capture data across a two-month span rather than a single month, specifically focusing on \textit{April Week 1--4} and \textit{May Week 1--4}. This entire evaluation window sits strictly behind the March 2026 corpus boundary to guarantee zero pre-training contamination.

\paragraph{Evaluation Settings} To bypass the operational latency associated with direct platform evaluation, we extract the historical future-past subset and align the timing parameters exactly with the original benchmark settings. We use $\Delta_T = \text{5 days}$ and establish the information freeze line as:
\begin{equation}
T_{\text{cut}} = T_{\text{end}} - \Delta_T = T_{\text{end}} - 5\text{ days}
\end{equation}

\paragraph{Metric} Performance on FutureX is measured multi-dimensionally across different question types (L1--L4) utilizing the original evaluation protocols:

\begin{itemize}
    \item \textbf{L1--L2 (Simple/multiple choice questions):} Evaluated using the standard \textbf{F1-score}, defined as:
    \begin{equation}
    \text{F1} = \text{F1}(Y,\hat{Y})=2 \times \frac{\text{Precision} \times \text{Recall}}{\text{Precision} + \text{Recall}}
    \end{equation}
    
    \item \textbf{L3 (Ranking):} For an ordered set of elements, given the ground-truth rank sequences $\{y_1, \dots, y_k\}$ and the predicted sequences $\{\hat{y}_1, \dots, \hat{y}_k\}$, the score is computed conditionally to award perfect ordering while penalizing partial match degradation:
    \begin{equation}
    \text{score}\left(\{y_1, \dots, y_k\}, \{\hat{y}_1, \dots, \hat{y}_k\}\right) = 
    \begin{cases} 
    1, & \text{if } y_i = \hat{y}_i, \text{ for } i = 1, \dots, k \\ 
    0.8 \times \frac{\left| \{y_1, \dots, y_k\} \cap \{\hat{y}_1, \dots, \hat{y}_k\} \right|}{k}, & \text{otherwise}
    \end{cases}
    \end{equation}

    \item \textbf{L4 (Numerical forecasting):} The alignment between a scalar ground truth $Y$ and its predicted continuous value $\hat{Y}$ is governed by a variance-normalized penalty function:
    \begin{equation}
    \text{score}(Y, \hat{Y}) = \max \left(0, \, 1 - \left(\frac{Y - \hat{Y}}{\sigma(Y)}\right)^2 \right)
    \end{equation}
    Since the true population standard deviation $\sigma(Y)$ cannot be directly extracted from the origin datasets, we approximate it by anchoring the scale relative to the ground-truth value, setting $\sigma(Y) = 1\% \times Y$. Therefore, the results of L4 are not suitable for direct comparison with the evaluation results of FutureX's open platform.
\end{itemize}

\subsection{Inference Settings and Context Management Strategies}
\label{appendix:inferencesettings}

\begin{table}[h]
\centering
\caption{Inference Parameter Settings for Student Models}
\label{tab:student_sampling_params}
\vspace{6px}
\small
\begin{tabular}{lc}
\hline
\textbf{Parameter} & \textbf{Value} \\ \hline
TOP\_P              & 0.95           \\
TEMP (Temperature)  & 1.0            \\
FREQUENCY\_PENALTY  & 0              \\
TOP\_K              & 20             \\
PRESENCE\_PENALTY   & 0              \\
REPETITION\_PENALTY & 1.0            \\ \hline
\end{tabular}
\end{table}

\paragraph{Configuration}
To evaluate the student models, we deploy the Supervised Fine-Tuning (SFT) variants using the vLLM framework \citep{vllm}. During inference, we utilize a sampling configuration detailed in Table~\ref{tab:student_sampling_params}. This sampling strategy introduces controlled stochasticity to diversify agent trajectories across independent runs, while simultaneously penalizing intra-trajectory repetitive outputs. To ensure alignment between training and inference distributions, we maintain the unified system prompt used in SFT data construction across both phases.

\begin{table}[h]
\centering
\caption{Context Management Settings for Student Models}
\label{tab:context_management_params}
\vspace{6px}
\small
\begin{tabular}{ll}
\hline
\textbf{Parameter} & \textbf{Value} \\ \hline
KEEP\_OBSERVATION\_THRESHOLD   & 3                     \\
MAX\_TOKEN\_LENGTH            & 32,000                \\
MAX\_ASSISTANT\_TURNS          & 100                   \\
MAX\_DUPLICATE\_TOOL\_CALL    & 3                     \\
MAX\_TOKEN\_PER\_ROUND        & 4,096                 \\
REPETITION\_PENALTY (Retry Limit) & 5                 \\ \hline
\end{tabular}
\end{table}

\paragraph{Context Management}
Complex Tool-Integrated Reasoning (TIR) tasks frequently induce structural or formatting anomalies, even in capable backbones such as Qwen3. To mitigate these formatting failures and enable a fair evaluation of core TIR capabilities, we implement a suite of runtime context-management mechanisms. The complete parameter specifications for context management are detailed in Table~\ref{tab:context_management_params}.

\begin{itemize}
    \item \textbf{Context Compression:} When the cumulative context length risks exhausting the processing window, we trigger an observation-level compaction strategy. Specifically, when the input prompt hits the sequence boundary, we retain only the 3 most recent environmental observations while older observation texts are systematically hidden.
    
    \item \textbf{Termination Conditions:} A trajectory is forcefully routed to a final summarization phase if it violates any of the conservative operational upper bounds: a maximum context length, a conversational turn limit, or a global tool-call repetition threshold (evaluated across the entire trajectory).
    
    \item \textbf{Force Summarize:} Upon hitting any termination condition, the system activates the force-summarize branch. The system appends the structured template suffix \texttt{<think>...</think><final\_summary>} to the end of the prompt, compelling the model to synthesize its final answer. Any subsequent tool execution requests are immediately purged to gracefully terminate the episode.
    
    \item \textbf{Exception Retry Mechanism:} If the model produces an anomalous response—defined as generating reasoning steps but failing to output either a valid tool call or a final summary—the system rejects the response and triggers a localized regeneration loop. This error-correction mechanism allows up to five retries per turn before the turn is abandoned.
\end{itemize}

\end{document}